\documentclass[10pt,twocolumn,letterpaper]{article}

\usepackage{cvpr}
\usepackage{times}
\usepackage{epsfig}
\usepackage{graphicx}
\usepackage{amsmath}
\usepackage{amssymb}
\usepackage{booktabs}       
\usepackage{multirow}
\usepackage{siunitx}
\usepackage{subcaption}
\usepackage{mathtools}
\usepackage{wrapfig}

\newcommand\blfootnote[1]{%
  \begingroup
  \renewcommand\thefootnote{}\footnote{#1}%
  \addtocounter{footnote}{-1}%
  \endgroup
}

\usepackage[pagebackref=true,breaklinks=true,letterpaper=true,colorlinks,bookmarks=false]{hyperref}

\cvprfinalcopy 

\def\cvprPaperID{5561} 

\ifcvprfinal\fi
\begin{document}

\title{Meta-learning Convolutional Neural Architectures for Multi-target Concrete Defect Classification with the COncrete DEfect BRidge IMage Dataset}

\author{
  Martin Mundt$^{1*}$, Sagnik Majumder$^{1}$, Sreenivas Murali$^{1*}$, Panagiotis Panetsos$^{2}$, Visvanathan Ramesh$^{1*}$ \\
  1. Goethe University $\quad$ 2. Egnatia Odos A. E. \\
  \tt\small{$\{$mmundt, vramesh$\}$@em.uni-frankfurt.de} $\quad$ {$\{$majumder, murali$\}$@ccc.cs.uni-frankfurt.de} \\ \tt\small{ppane@egnatia.gr} \\
}

\maketitle

\begin{abstract}
\blfootnote{* work conducted while at Frankfurt Institute for Advanced Studies}  Recognition of defects in concrete infrastructure, especially in bridges, is a costly and time consuming crucial first step in the assessment of the structural integrity. Large variation in appearance of the concrete material, changing illumination and weather conditions, a variety of possible surface markings as well as the possibility for different types of defects to overlap, make it a challenging real-world task. 
In this work we introduce the novel COncrete DEfect BRidge IMage dataset (CODEBRIM) for multi-target classification of five commonly appearing concrete defects. We investigate and compare two reinforcement learning based meta-learning approaches, MetaQNN and efficient neural architecture search, to find suitable convolutional neural network architectures for this challenging multi-class multi-target task. We show that learned architectures have fewer overall parameters in addition to yielding better multi-target accuracy in comparison to popular neural architectures from the literature evaluated in the context of our application. 
\end{abstract}

\section{Introduction}
To assess a concrete bridge's structural safety, it is desirable to determine the level of degradation by accurately recognizing all defect types. Defects tend to be small with respect to bridge elements and often occur simultaneously with overlap of defect categories. Although one could imagine treating each defect category independently, overlapping defects are more severe with respect to the structural safety. The requirement to recognize these multi-class multi-target defects forms the basis for a challenging real-world task that is further complicated by a variety of environmental factors. Concrete, as a composite material, has a wide range of variation in surface reflectance, roughness, color and, in some cases, applied surface coatings. Changing lighting conditions, weather dependent wetness of the surface and a diverse set of safety irrelevant surface alterations like small holes, markings, stains or graffiti, add to the factors of variation. This necessitates computer vision techniques that are capable of addressing such rich appearance spaces.  

Deep learning techniques in conjunction with labelled datasets have turned out to be ideal candidates for recognition tasks of similar complexity. Especially convolutional neural networks (CNNs) \cite{Krizhevsky2012, Simonyan2015, Andrearczyk2016, Zagoruyko2016, Huang2016} have been shown to excel at object and material recognition benchmarks \cite{ILSVRC, Everingham2015, Xiao2010, Bell2015}. Unfortunately, defect recognition in concrete bridges is largely yet to benefit from deep learning approaches. Due to the necessity of expert knowledge in the annotation process along with tedious image acquisition, the task is traditionally focused on cracks with algorithms based on domain specific modelling or manual inspection by a human.
Recently, datasets \cite{Shi2016, Yang2017, Macquire2018} and corresponding deep learning applications \cite{Yang2017, Li2018, Kim2018, DaSilva2018} have presented significant efforts towards data-driven approaches in this domain. Their work focuses on cracks as only a subset of structurally relevant defects and concentrates on CNNs proposed in the object recognition literature, that might not be the best choice for material defect recognition.

In this work we address two crucial open aspects of concrete defect recognition: the establishment of a labelled multi-target dataset with overlapping defect categories for use in machine learning and the design of deep neural networks that are tailored to the task. For this purpose we introduce our novel COncrete DEfect BRidge IMage (CODEBRIM) dataset and employ meta-learning of CNN architectures specific to multi-class multi-target defect classification. 
CODEBRIM features six mutually non-exclusive classes: crack, spallation, efflorescence, exposed bars, corrosion (stains) and non-defective background. Our images were acquired at high-resolution, partially using an unmanned aerial vehicle (UAV) to gain close-range access, and feature varying scale and context. We evaluate a variety of best-practice CNN architectures \cite{Krizhevsky2012, Simonyan2015, Andrearczyk2016, Zagoruyko2016, Huang2016} in the literature on the CODEBRIM's multi-target defect recognition task. We show that meta-learned neural architectures achieve equivalent or better accuracies, while being more parameter efficient, by investigating and comparing two reinforcement learning neural architecture search approaches: MetaQNN \cite{Baker2016} and "efficient neural architecture search" (ENAS) \cite{Pham2018}. The CODEBRIM dataset is publicly available at: https://doi.org/10.5281/zenodo.2620293 . We also make the code for training the CNN baselines and both meta-learning techniques available open-source at: https://github.com/MrtnMndt/meta-learning-CODEBRIM . To summarize our contributions:

\begin{itemize}
	\item We introduce a novel high-resolution multi-class multi-target dataset featuring images of defects in context of concrete bridges. 
	\item We evaluate and compare best-practice CNN architectures for the task of multi-target defect classification. 
	\item We adapt and contrast two reinforcement learning based architecture search methods, MetaQNN and ENAS, on our multi-target scenario. We show how resulting meta-learned architectures from both methods improve the presented task in terms of higher accuracy and lower model parameter count.  
\end{itemize}

\section{Prior and related work}
\paragraph*{Datasets.}
Image classification and object detection benchmarks predominantly focus on the single-target scenario. Popular examples are the ImageNet \cite{ILSVRC}, Pascal VOC \cite{Everingham2015} or the scene understanding SUN dataset \cite{Xiao2010}, where the task is to assign a specific class to an image, area or pixel. Much of the recent computer vision deep learning research is built upon improvements based on these publicly available datasets. The "materials in context" database (MINC) \cite{Bell2015} followed in spirit and has created a dataset for material and texture related recognition tasks. To a large degree MINC has extended previous datasets and applications built upon prior work of the (CUReT) database \cite{Dana1999}, the FMD dataset \cite{Sharan2009} and KTH-TIPS \cite{Hayman2004, Caputo2005}. With respect to defects in concrete structures, or bridges in particular, openly available datasets remain scarce. Depending on the defect type that needs to be recognized, our task combines texture anomalies such as efflorescence or cracks with objects such as exposed reinforcement bars. Domain specific dataset contributions were very recently proposed with the "CrackForest" dataset \cite{Shi2016}, the CSSC database \cite{Yang2017} and SDNET2018 \cite{Macquire2018}. However, as all of the former works feature a single-target and in fact single-class task, we have decided to extend existing work with the multi-class multi-target CODEBRIM dataset. 

\paragraph*{Defect (crack) recognition.}
Koch \etal \cite{Koch2015} provide a comprehensive review on the state of computer vision in concrete defect detection and open aspects. In summary, the majority of approaches follow task specific modelling. Data-driven applications are still the exception and are yet to be leveraged fully. Recent works  \cite{Li2018, DaSilva2018, Kim2018} show application to crack versus non-crack classification using images with little clutter and lack of structural context. An additional defect class of spalling is considered by the authors of \cite{Yang2017}. Similar to other works, they focus on the single-target scenario and evaluation of well-known CNN baselines from prior object recognition literature. We extend their work by meta-learning more task specific neural architectures for more defect categories and overlapping defects.  

\paragraph*{Convolutional neural networks.}
A broader review of deep learning, its history and neural architecture innovations is given by LeCun \etal \cite{Lecun2015}. We recall some CNN architectures that serve as baselines and give a frame of reference for architectures produced by meta-learning on our task. Alexnet \cite{Krizhevsky2012} had a large success on the ImageNet \cite{ILSVRC} challenge that was later followed by a set of deeper architectures commonly referred to as VGG \cite{Simonyan2015}. Texture-CNN \cite{Andrearczyk2016} is an adapted version of the Alexnet design that includes an energy-based adaptive feature pooling and FV-CNN \cite{Cimpoi2015} augments VGG with Fisher Vector pooling for texture classification. Recent works address information flow in deeper networks by adding skip connections with residual networks \cite{He2016}, wide residual networks (WRN) \cite{Zagoruyko2016} and densely connected networks (DenseNet) \cite{Huang2016}. 

\paragraph*{Meta-learning neural architectures.}
Although deep neural networks empirically work well in many practical applications, networks have initially been designed for different tasks. A recent trend to bypass the human design intuition is to treat neural architectures themselves from a meta-learning perspective and conduct a black-box optimization on top of the training of weights to find suitable task-specific architecture designs. 
Several works in the literature have posed architecture meta-learning from a variety of perspectives based on reinforcement learning (RL) controllers \cite{Baker2016, Zoph2017, Pham2018, Cai2018}, differentiable methods \cite{Liu2018} or evolutionary strategies \cite{Real2017}. In our work, we evaluate and adapt two RL based approaches to multi-target defect classification: MetaQNN \cite{Baker2016} and "efficient neural architecture search" (ENAS) \cite{Pham2018}. We pick these two approaches as they share underlying principles of training RL controllers. This allows us to pick a common reward metric determined by proposed CNN candidate accuracies. The main differences lie in the RL agents' nature: MetaQNN employs Q-Learning to learn to suggest increasingly accurate CNNs, whereas ENAS uses policy gradients \cite{Sutton1999} to train an auto-regressive recurrent neural network that samples individual layers based on previous input.

\begin{figure*}[t!]
	\centering
	\begin{subfigure}{\textwidth}
		\includegraphics[width= 0.245 \textwidth]{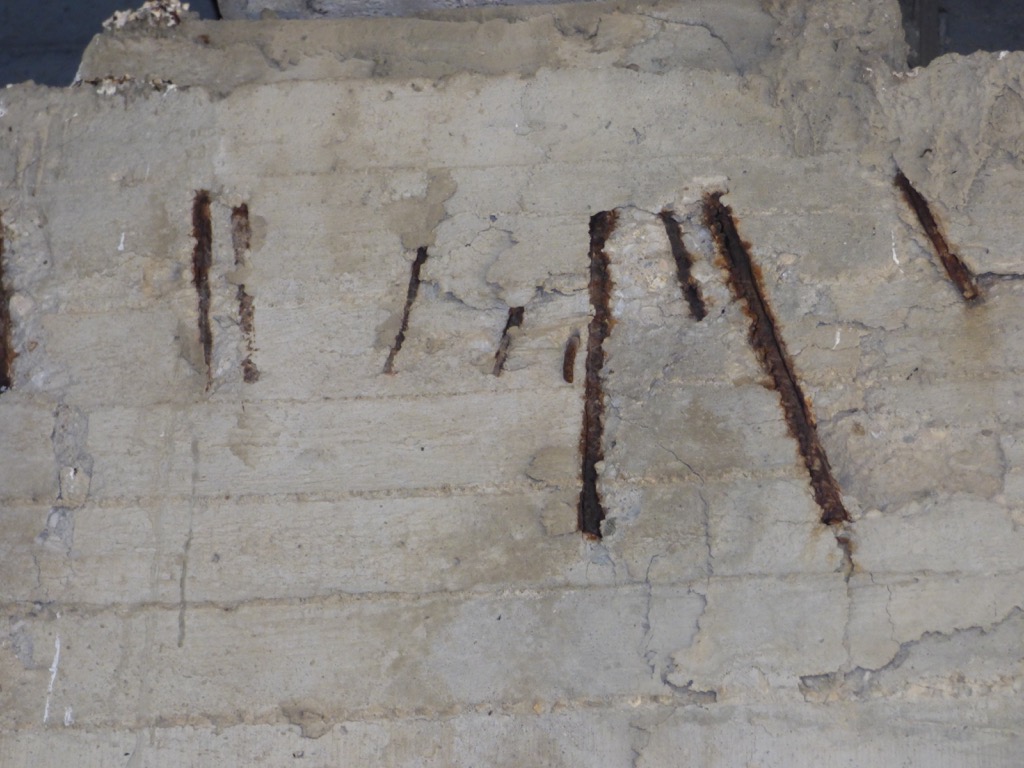}
		\includegraphics[width= 0.245 \textwidth]{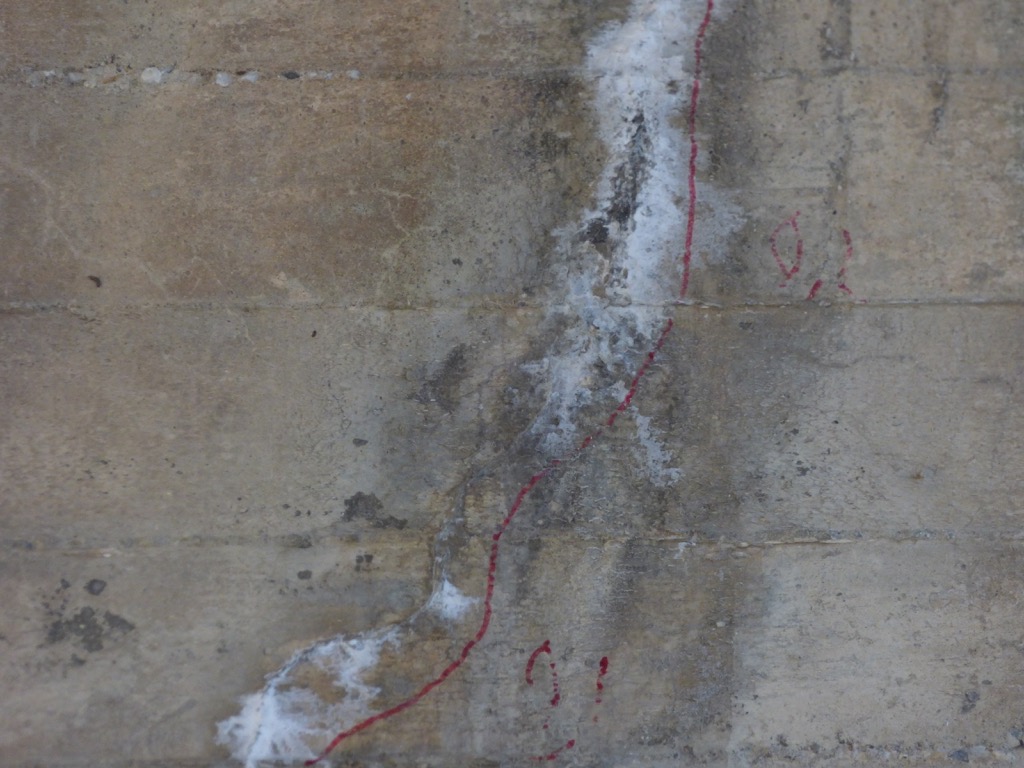}
		\includegraphics[width= 0.245 \textwidth]{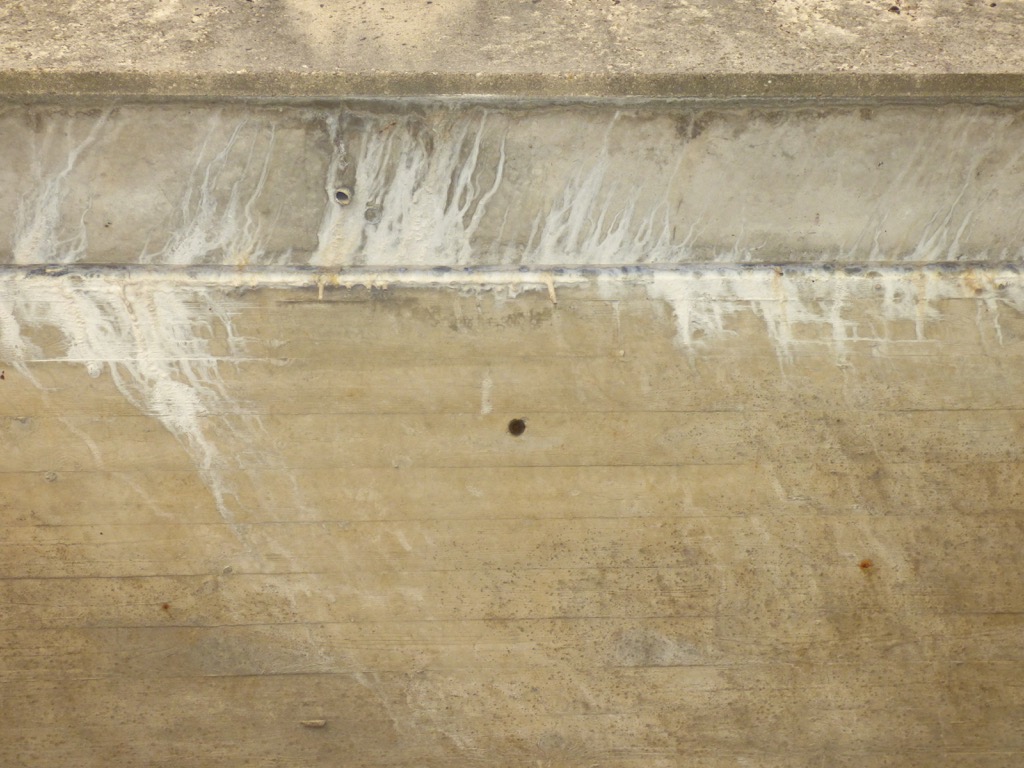}
		\includegraphics[width= 0.245 \textwidth]{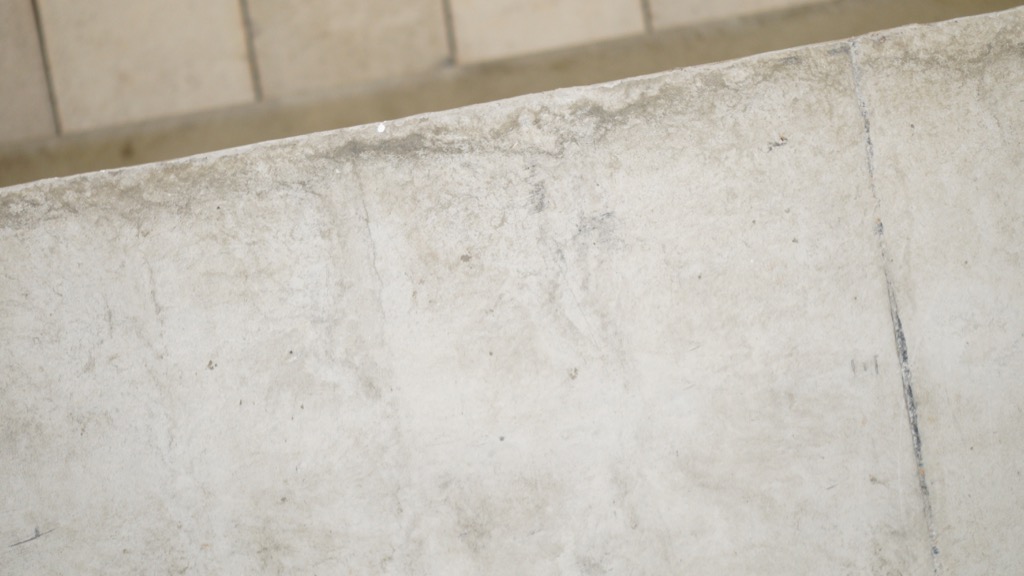} \\
		\smallskip
		\includegraphics[width= 0.245 \textwidth]{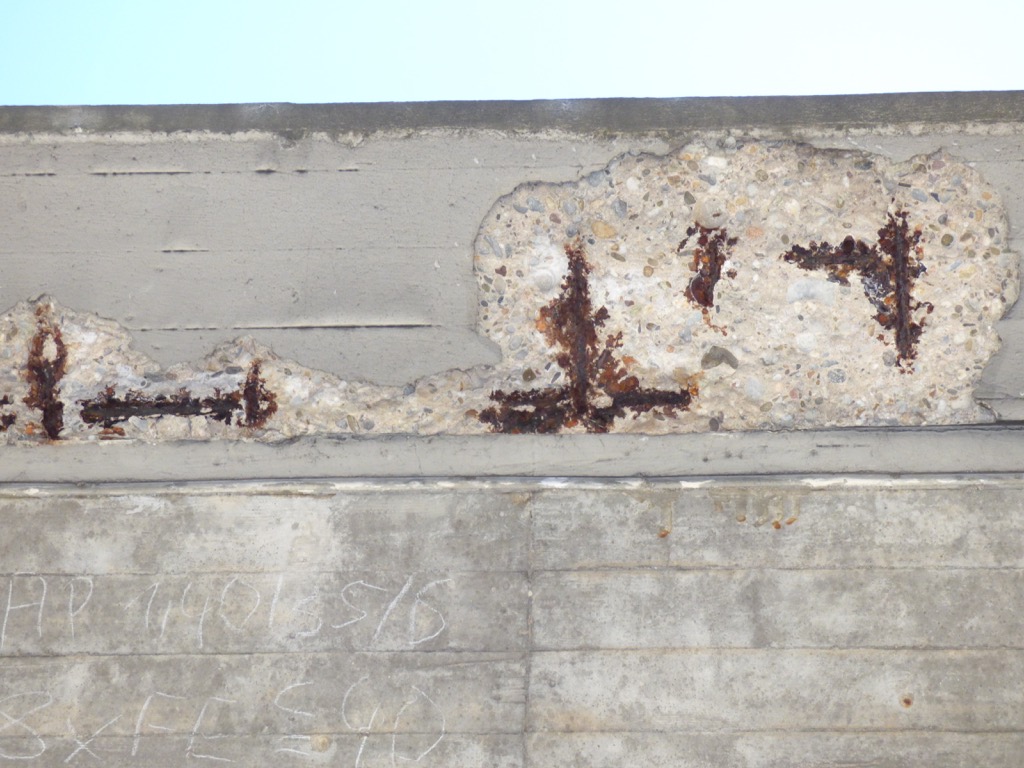} 
		\includegraphics[width= 0.245 \textwidth]{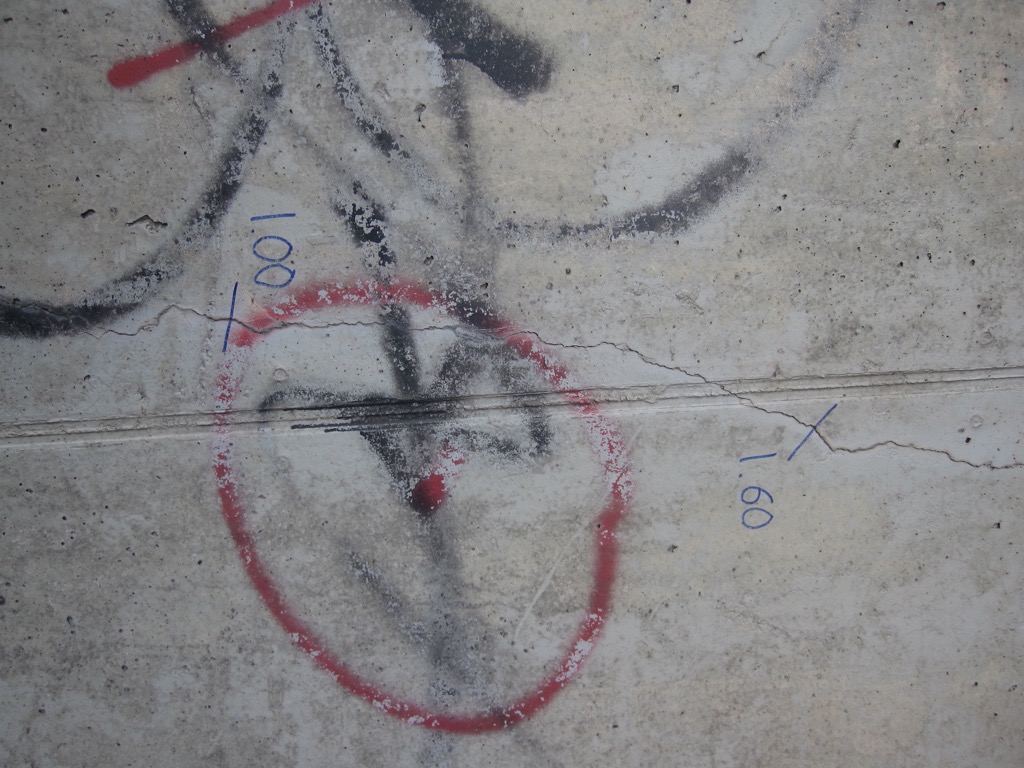} 
		\includegraphics[width= 0.245 \textwidth]{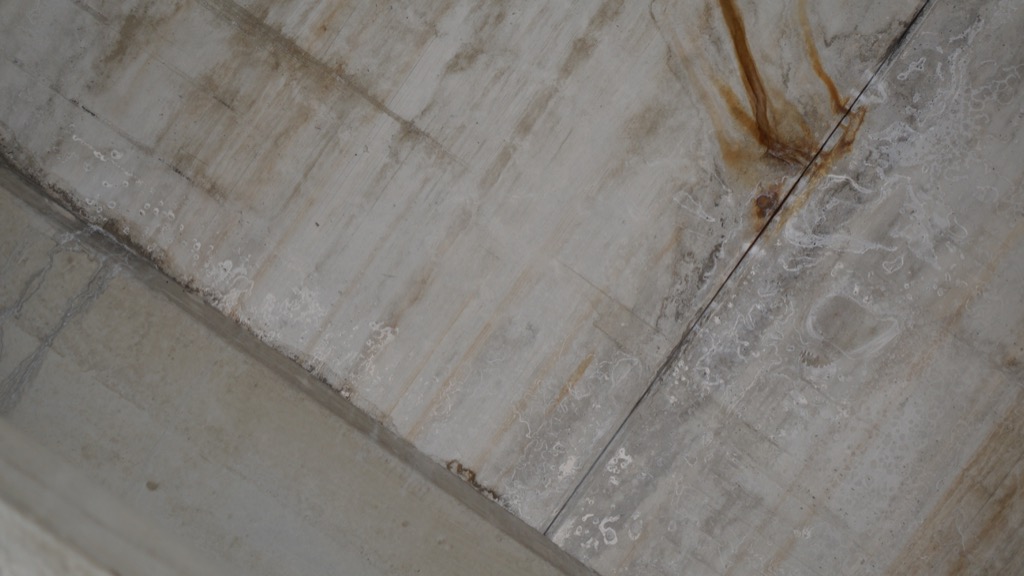} 
		\includegraphics[width= 0.245 \textwidth]{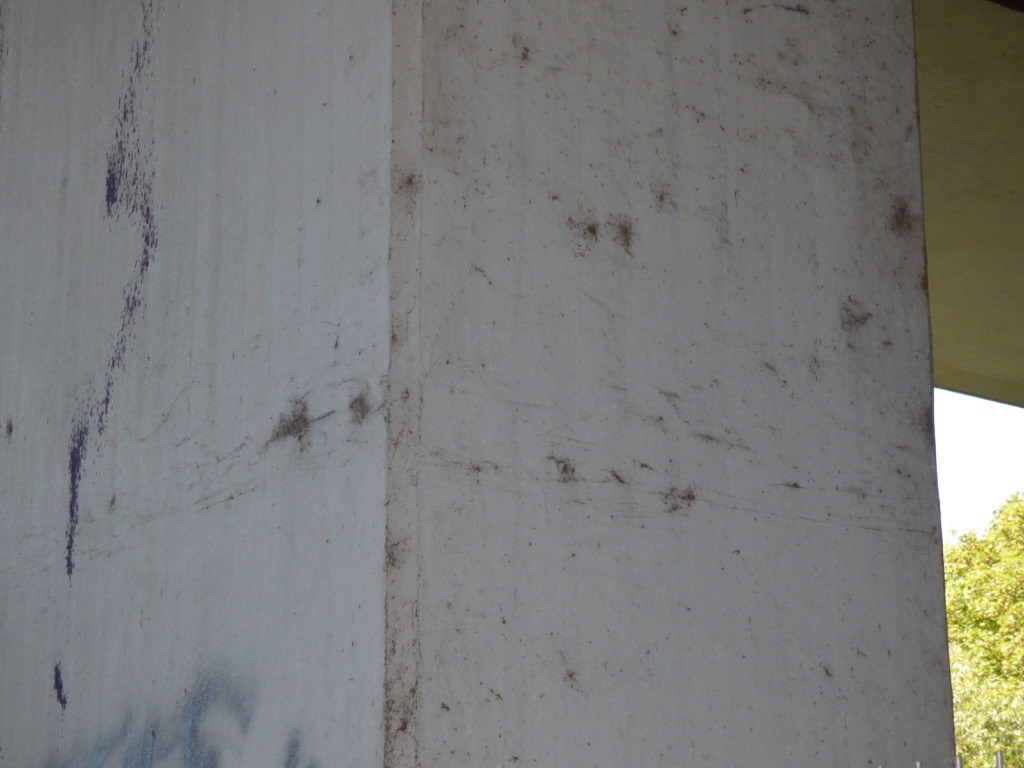} 
		\subcaption{Top row from left to right: 1.) exposed bars, spallation, cracks (hard to see) 2.) hairline crack with efflorescence 3.) efflorescence 4.) defect-free concrete. Bottom row from left to right: 1.) large spalled area with exposed bars and corrosion 2.) crack with graffiti 3.) corrosion stain, minor onset efflorescence 4.) defect-free concrete with dirt and markings.}
		\label{fig:dataset_examples_full}
	\end{subfigure}
	\centering
	\begin{subfigure}{\textwidth}
		\includegraphics[width= 0.1 \textwidth]{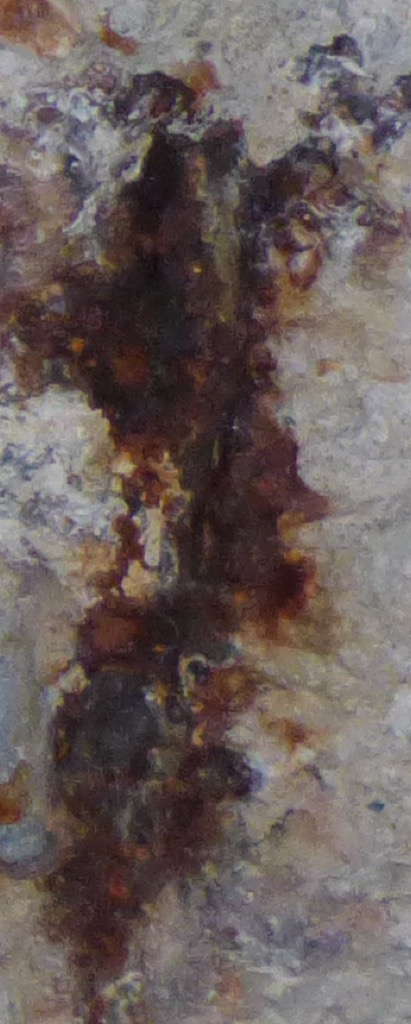}
		\includegraphics[width= 0.05 \textwidth]{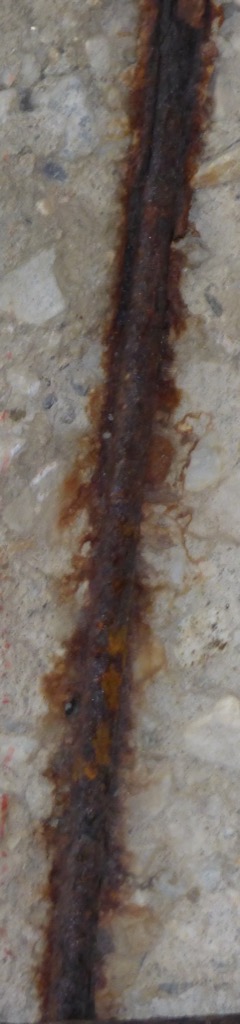}
		\includegraphics[width= 0.05 \textwidth]{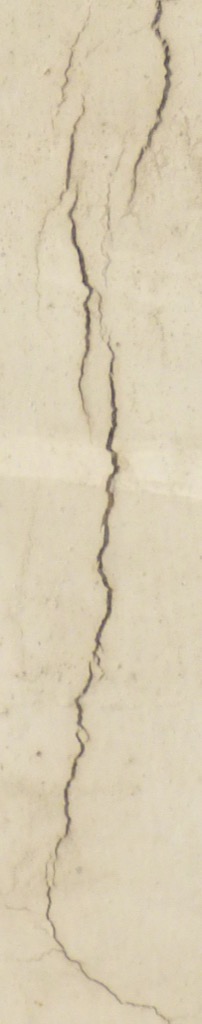}
		\includegraphics[width= 0.1 \textwidth]{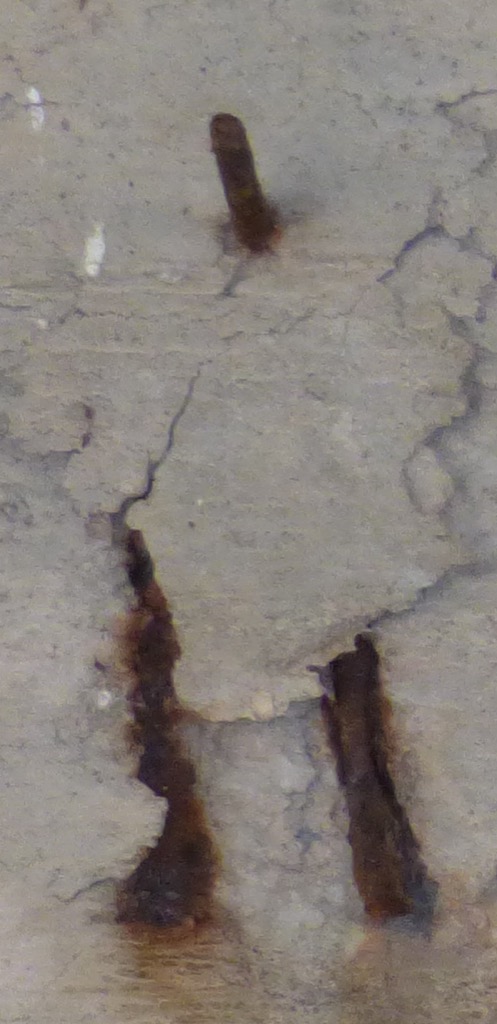}
		\includegraphics[width= 0.045 \textwidth]{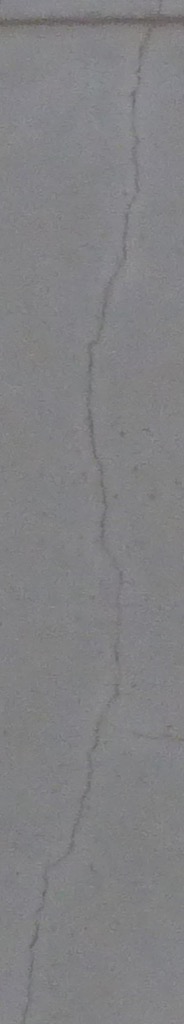}
		\includegraphics[width= 0.1 \textwidth]{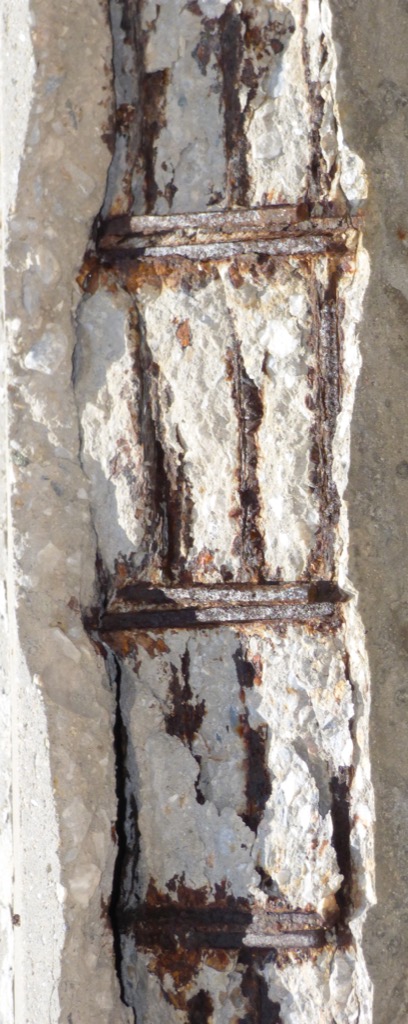} 
		\includegraphics[width= 0.15 \textwidth]{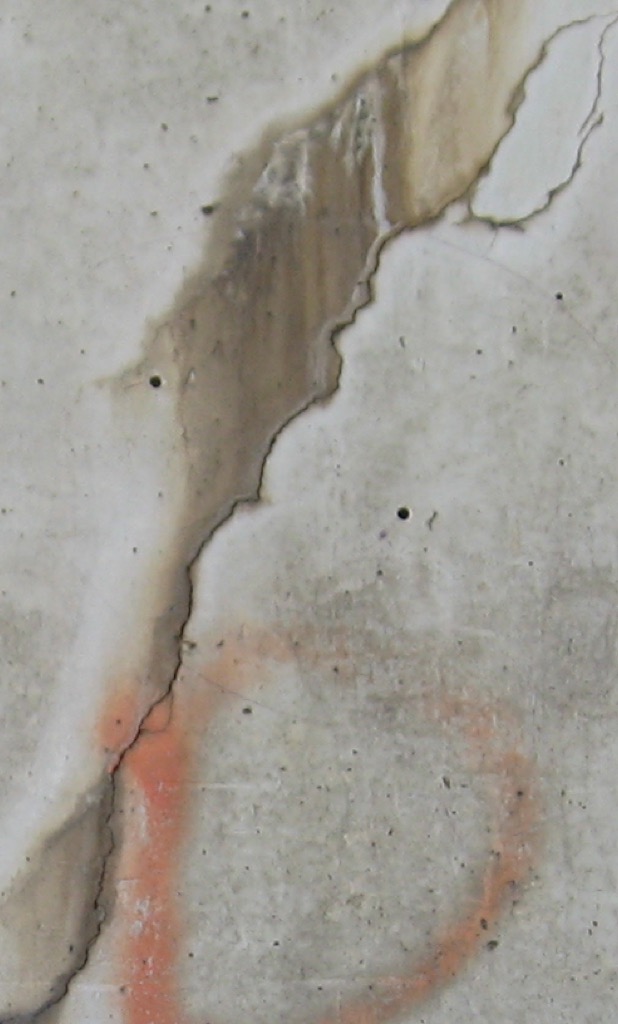} 
		\includegraphics[width= 0.09 \textwidth]{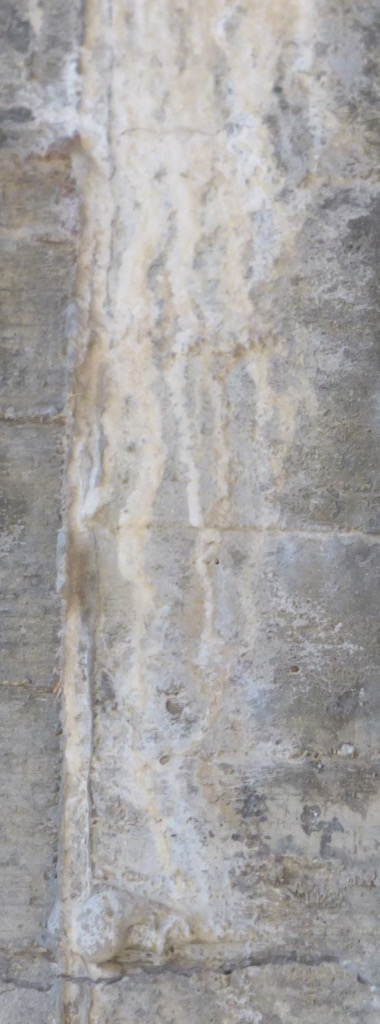} 
		\includegraphics[width= 0.15 \textwidth]{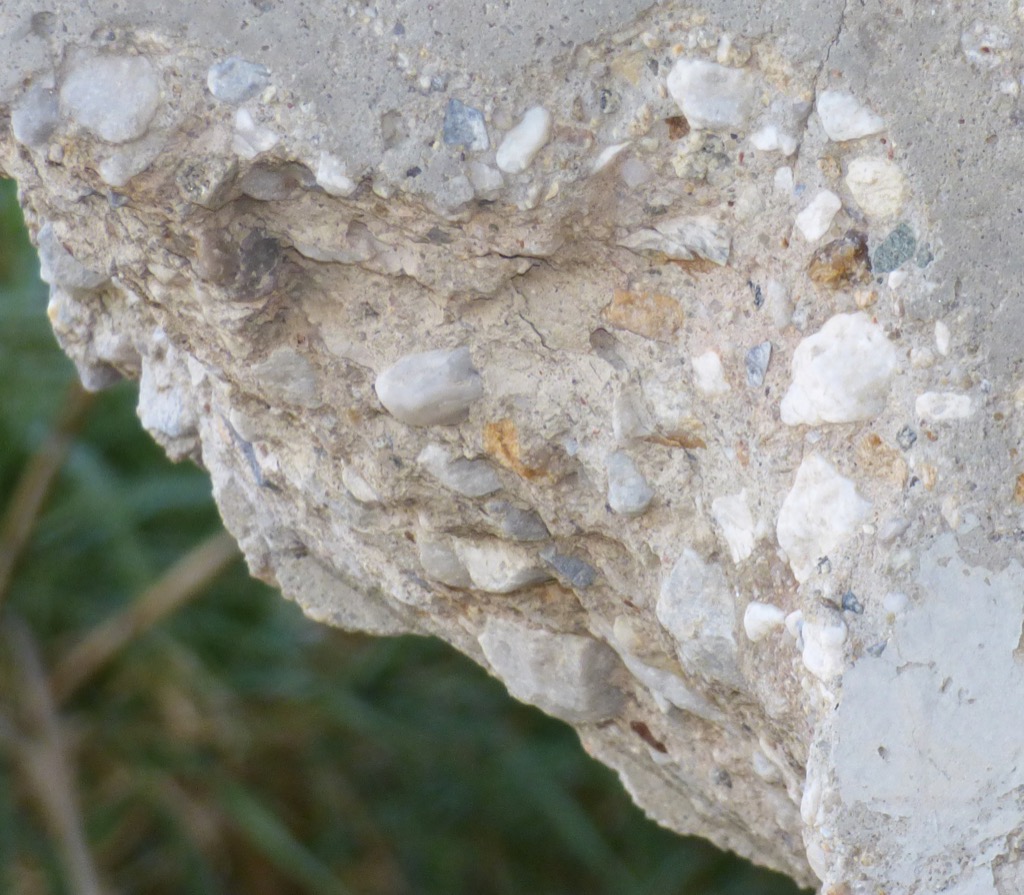} 
		\includegraphics[width= 0.12 \textwidth]{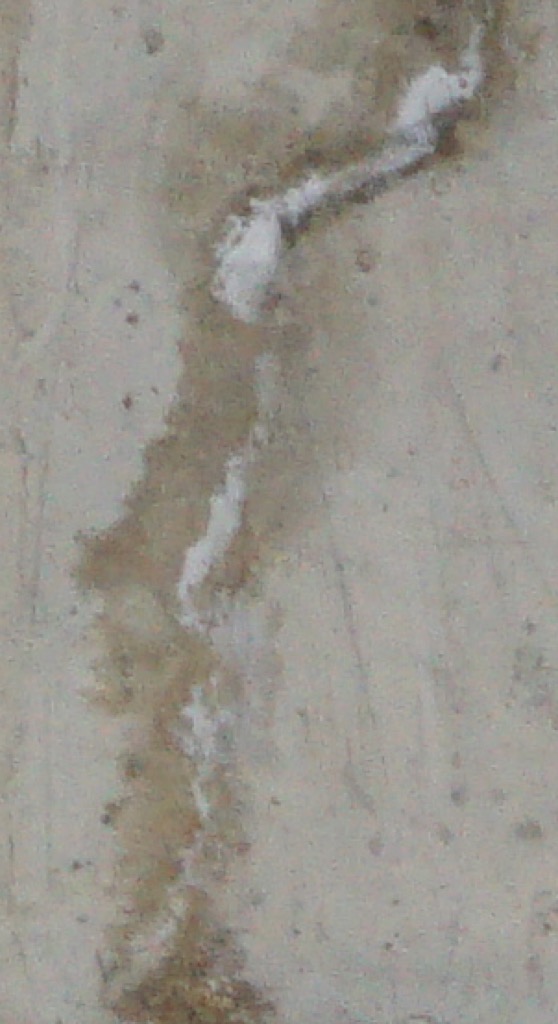}
		\caption{From left to right: 1.) spalled area with exposed bar, advanced corrosion and efflorescence 2.) exposed corroded bar 3.) larger crack 4.) partially exposed corroded bars, cracks 5.) hairline crack 6.) heavy spallation, exposed bars, corrosion 7.) wet/damp crack with efflorescence on the top 8.) efflorescence 9.) spalled area 10.) hairline crack with efflorescence.}
		\label{fig:dataset_examples_crops}
	\end{subfigure}
	\caption{Dataset examples. Top figure: full high-resolution images. Images heavily down-sampled for view in pdf. Bottom figure: Image patches cropped from annotated bounding boxes (not corresponding to top images). Images resized for view in pdf but with original aspect ratio.}
	\label{fig:dataset_examples}
\end{figure*}

\section{The CODEBRIM dataset}
The acquisition of the COncrete DEfect BRidge IMage: CODEBRIM dataset was driven by the need for a more diverse set of the often overlapping defect classes in contrast to previous crack focused work \cite{Shi2016, Yang2017, Macquire2018}. In particular, deep learning application to a real-world inspection scenario requires sampling of real-world context due to the many factors of variation in visual defect appearance. Our dataset is composed of five common defect categories: crack, spallation, exposed reinforcement bar, efflorescence (calcium leaching), corrosion (stains), found in 30 unique bridges (disregarding bridges that did not have defects). The bridges were chosen according to varying overall deterioration, defect extent, severity and surface appearance (e.g. roughness and color). Images were taken under changing weather conditions to include wet/stained surfaces with multiple cameras at varying scales. As most defects tend to be very small one crucial requirement was the acquisition at high-resolution. Considering that large parts of bridges are not accessible for a human, a subset of our dataset was acquired by UAV. We continue with the requirements and rationale behind the camera choices, the annotation process that led to the dataset and finally give a summary of important dataset properties. 

\subsection{Image acquisition and camera choice}
Image acquisition and camera choices were motivated by typical concrete cracks in bridges having widths as small as $\SI{0.3}{\milli\metre}$ \cite{Koch2015}. Resolving such defects on a pixel level imposes a strong constraint on the distance and resolution at which the images are acquired. In a naive calculation for a conventional consumer-grade camera with an example chip of size $23.50 \times \SI{15.60}{\milli\metre}$ and maximum resolution $6000 \times 4000$, this translates to around $\SI{0.1}{\milli\metre}$ per pixel at a focal length of $\SI{50}{\milli\metre}$ and a distance of roughly $\SI{1.5}{\metre}$ (assuming a pinhole camera model and viewing axis perpendicular to the surface). Based on this requirement our dataset was gathered with four different cameras at high resolution and large focal lengths under varying distance and angles. In addition, to homogeneously illuminate the darker bridge areas, we made use of diffused flash. Exact camera models and corresponding detailed parameters can be found in the supplementary material.

\subsection{Dataset properties}
\begin{figure}[t]
	\centering
	\includegraphics[width= \columnwidth]{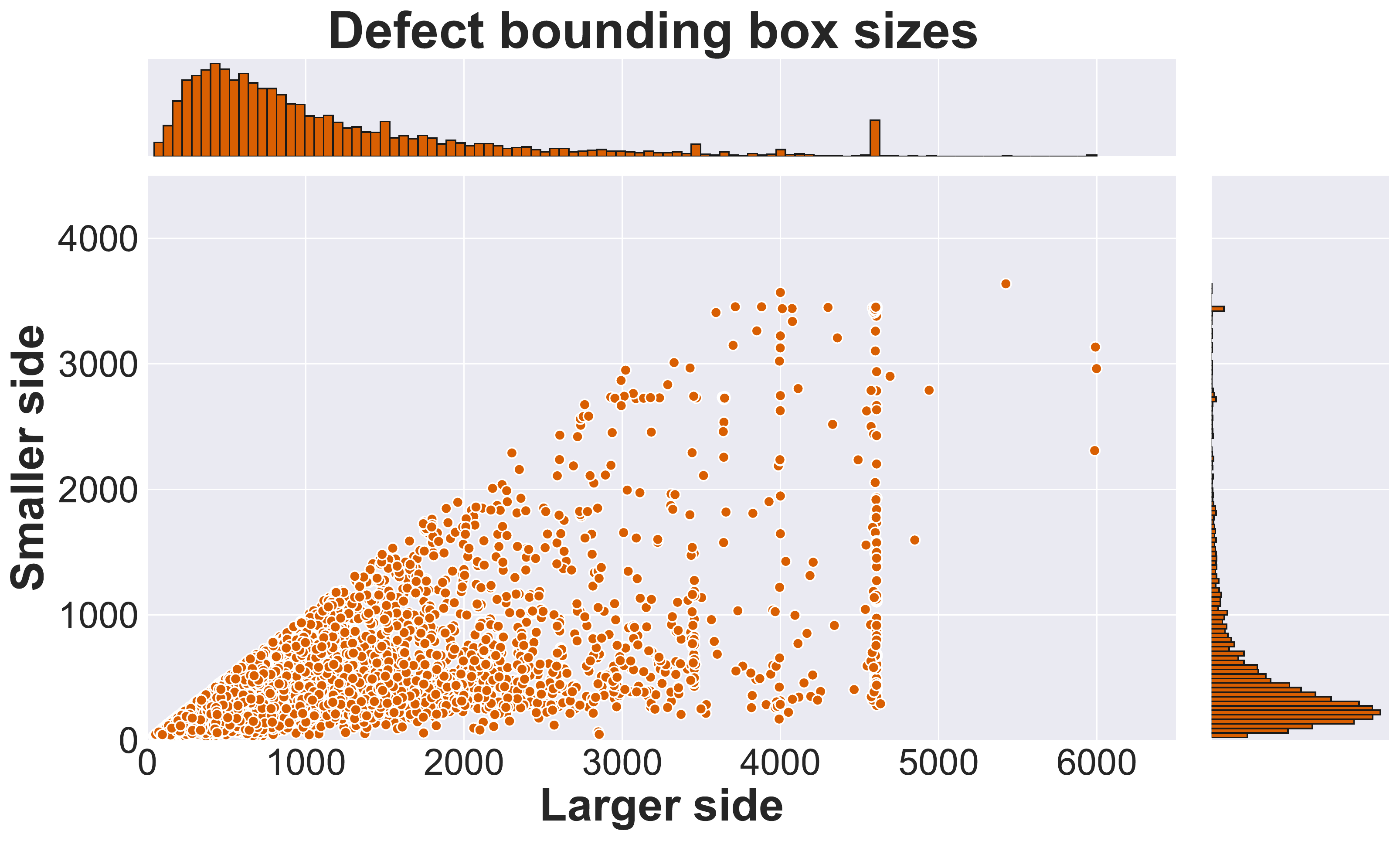} \\
	\includegraphics[width= \columnwidth]{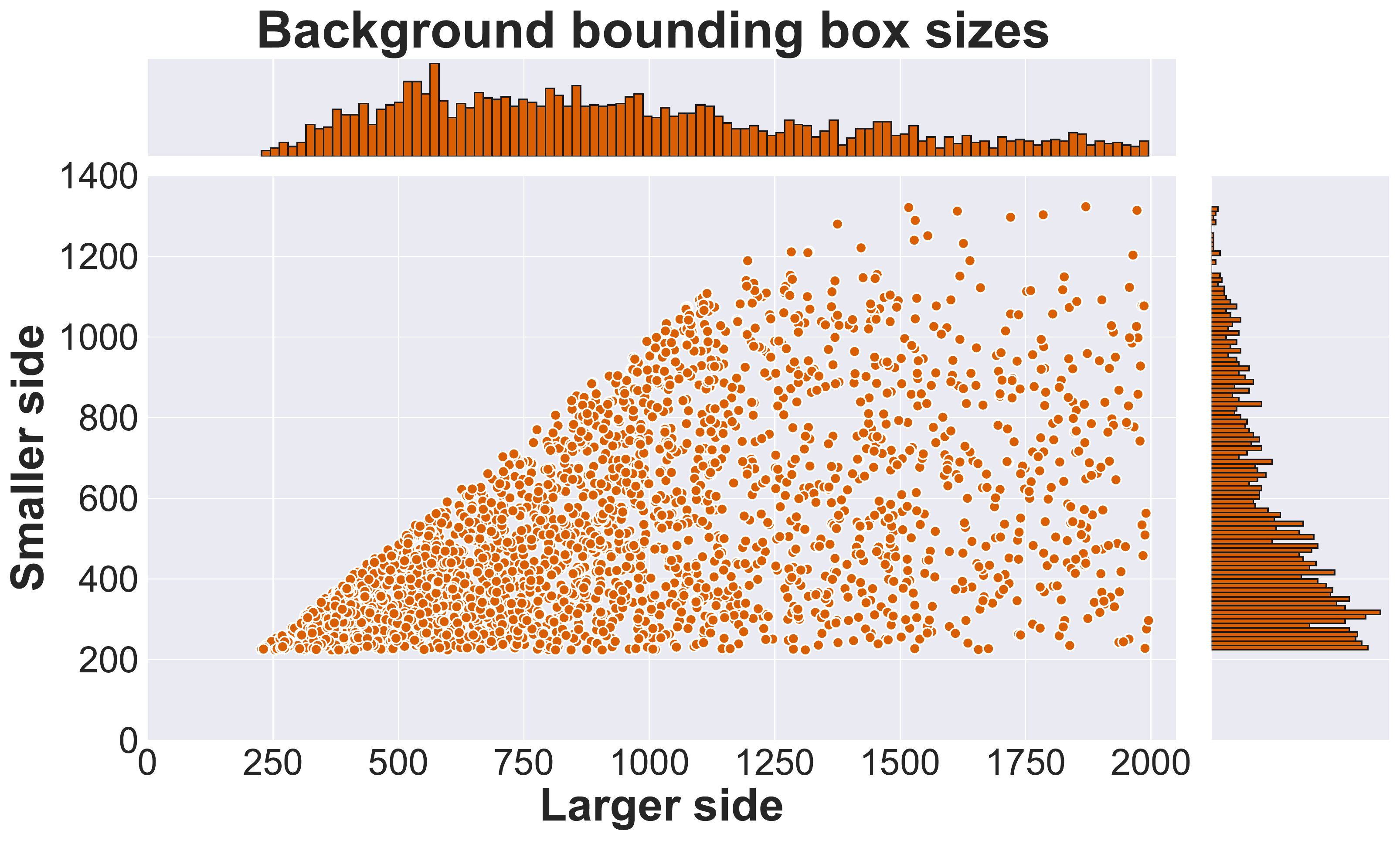}
	\caption{Top panel: distribution of annotated bounding box sizes for defects. Bottom panel: distribution of sizes for sampled non-overlapping background bounding boxes.}
	\label{fig:sizes}
\end{figure}
We employed a multi-stage annotation process by first curating acquired images, annotating bounding boxes per defect and sequentially labelling each class separately. The rationale and exact annotation process is outlined in the supplementary material. The acquisition and annotation process resulted in a dataset with the following properties:

\begin{itemize}
\item 1590 high-resolution images with defects in context of 30 unique bridges, acquired at different scales and resolutions.
\item 5354 annotated defect bounding boxes (largely with overlapping defects) and 2506 generated non-overlapping background bounding boxes.
\item Defect numbers for the following classes: crack - 2507, spallation - 1898, efflorescence - 833, exposed bars - 1507 and corrosion stain - 1559.
\end{itemize}

Examples of images and extracted patches from bounding boxes featuring a variety of overlapping and non-overlapping defects can be seen in figure \ref{fig:dataset_examples_full} and \ref{fig:dataset_examples_crops} respectively. We point out that in contrast to most object and texture based benchmarks, the majority of our dataset has more than one class occurring at once. We show a corresponding histogram for the number of defect classes per individual bounding box annotation in the supplementary material. 

Apart from the multi-target nature making our dataset more challenging than single-class recognition, the task is difficult because of large variations in aspect ratio, scale and resolution of the different defects and their bounding boxes. This is true especially at a scene level, considering that cracks can be very fine and elongated, whereas spalled areas can vary almost arbitrarily. To illustrate these variations we visualize the distributions of defect bounding box sizes and the sampled background bounding box sizes in figure \ref{fig:sizes}. Further details about distributions of image sizes, bounding box size distributions per category (with overlaps due to the multi-target nature) and distribution of aspect ratios per defect can be found in the supplementary material. 

\section{Meta-learning convolutional neural networks for multi-target defect classification}
\begin{table*}[t]
\resizebox{\textwidth}{!}{\begin{tabular}{llrrrrrrrrr}
  &  & \multicolumn{9}{c}{Multi-target accuracy [\%] depending on learning rate schedule: max to min}  \\ 
 \cmidrule{3-11}
 Architecture & Batch size & \multicolumn{3}{c}{$\left[10^{-1}, 10^{-5}\right]$} & \multicolumn{3}{c}{$\left[5 \cdot 10^{-2}, 5 \cdot 10^{-4} \right]$} & \multicolumn{3}{c}{$\left[10^{-2}, 10^{-5}\right]$} \\ 
  &  & best val & bv-test & bv-train & best val & bv-test & bv-train & best val & bv-test & bv-train  \\ 
\toprule
\multirow{4}{*}{T-CNN} & 16 & 64.62 & 69.51 & 80.27 & 63.67 & 65.71 & 83.38 & 64.30 & 67.93 & 93.91  \\ 
 & 32 & 64.78 & 66.19 & 87.66 & 63.36 & 68.72 & 94.49 & 62.84 & 66.35 & 96.22   \\ 
 & 64 & 63.36 & 70.14 & 95.21 & 63.52 & 67.93 & 98.10 & 62.26 & 66.82 & 95.85   \\ 
 & 128 & 63.67 & 67.45 & 98.31 & 63.36 & 66.82 & 98.63 & 60.53 & 65.08 & 94.47   \\ 
\midrule
\multirow{4}{*}{VGG-A} & 16 & 60.22 & 62.08 & 75.74 & 63.67 & 68.24 & 94.78 & 64.93 & 70.45 & 98.29   \\ 
 & 32 & 63.05 & 67.77 & 93.88 & 63.05 & 66.35 & 94.27 & 65.40 & 69.51 & 97.01 \\ 
 & 64 & 63.36 & 69.66 & 98.00 & 63.37 & 70.45 & 90.64 & 59.90 & 63.82 & 97.01  \\ 
 & 128 & 63.20 & 61.29 & 92.99 & 63.52 & 68.07 & 98.55 & 58.80 & 61.29 & 92.99   \\ 
\bottomrule
\end{tabular}}
\caption{Grid-search conducted on different batch sizes and different learning rate schedules for the T-CNN and VGG-A models. The multi-target best validation accuracy (best val) is shown together with each model's accuracy on the test set at the point in time of achieving the best validation accuracy (bv-test). The analogous training accuracy (bv-train) is shown to demonstrate that models do not under-fit. These validation accuracies have been used to determine training hyper-parameters.}
\label{tab:hyperparams_selection}
\end{table*}

We use meta-learning to discover models tailored to multi-target defect classification on the CODEBRIM dataset. In order to find a suitable set of hyper-parameters for the meta-learning search space and training of neural architectures we start with the T-CNN \cite{Andrearczyk2016} and VGG-A \cite{Simonyan2015} baselines and investigate the influence of learning rate, batch size and patch size. For this we separate the dataset into train and validation splits and set aside a final test set for evaluation. We then adapt the MetaQNN \cite{Baker2016} and ENAS \cite{Pham2018} architecture meta-learning approaches and contrast the obtained results with the following set of CNN architectures proposed in the literature: Alexnet \cite{Krizhevsky2012}, T-CNN \cite{Andrearczyk2016}, VGG-A and VGG-D \cite{Simonyan2015}, wide residual network (WRN) \cite{Zagoruyko2016} and densely connected convolutional networks (DenseNet) \cite{Huang2016}. We want to point out that even though bounding box annotations are present in our dataset, we do not evaluate any bounding box detection algorithms because our goal at this stage is the establishment of the already challenging multi-target classification task. We have also evaluated transfer-learning from the ImageNet and MINC datasets, albeit without improvements and therefore report these experiments in the supplementary material. 

\subsection{Dataset training, validation and test splits}
We have randomly chosen 150 unique defect examples per class for validation and test sets respectively. To avoid over-fitting due to very similar context, we make sure that we always include all annotated bounding boxes from one image in one part of the dataset split only. An alternative way to split the dataset is to separate train, validation and test sets according to unique bridges. However, it is infeasible to balance such a split with respect to equal amount of occurrences per defect due to individual bridges not featuring defect classes uniformly (particularly with class overlaps) and thus makes an unbiased training and reporting of average losses or accuracies difficult. Nevertheless, to investigate the importance of over-fitting global properties, we investigate and further discuss the challenges of such splits in the supplementary material. 

\subsection{Training procedure}
The challenging multi-class multi-target nature of our dataset makes the following measures necessary:
\begin{enumerate}
\item \textbf{Multi-class multi-target.} For a precise estimate of a model's performance in a multi-target scenario, a classification is considered as correct if, and only if, all the targets are predicted correctly. To adapt all neural networks for this scenario we use a Sigmoid function for every class in conjunction with the binary cross entropy loss function. When we calculate classification accuracies we binarize the Sigmoid output with a threshold of $0.5$. Note that this could be treated as a hyper-parameter to potentially obtain better results.
\item \textbf{Variations in scale and resolution.} We address the variation in scale and resolution of bounding boxes by following the common literature approach based on previous datasets such as ImageNet \cite{ILSVRC} and the models presented in \cite{Krizhevsky2012, Simonyan2015, Zagoruyko2016, Huang2016}. Here, the smaller side of the extracted patch is rescaled to a pre-determined patch size and random quadratic crops of patch size are taken to extract fixed size images during training.
\item \textbf{Train set imbalance.} We balance the training dataset by virtually replicating the under-represented class examples such that the overall defect number per class is on the same scale to make sure defect classes are sampled equally during training. Note that test and validation sets are balanced by design.
\end{enumerate}

The reason for adopting step two is to allow for a direct comparison with CNNs proposed in prior literature without making modifications to their architectures. We do not use individual class accuracies as a performance metric as it is difficult to compare models that don't capture overlaps adequately. Nevertheless we provide an example table with multi-target versus per-class accuracy of later shown CNN literature baselines in the supplementary material. 

\subsubsection{Common hyper-parameters}
We conduct an initial grid-search to find a suitable common set of hyper-parameters for CNNs (meta-learned or not) trained with stochastic gradient descent based on the T-CNN \cite{Andrearczyk2016} and VGG-A \cite{Simonyan2015} architectures. For this we use learning rate schedules with warm restarts (SGDWR) according to the work of \cite{Loshchilov2017}. The grid search features three cycles with ranges inspired by previous work \cite{Loshchilov2017, Pham2018}: $\left[10^{-1}, 10^{-5}\right], \left[5 \cdot 10^{-2}, 5 \cdot 10^{-4} \right]$ and $\left[10^{-2}, 10^{-5}\right]$, a warm restart cycle length of 10 epochs that is doubled after every restart, and four different batch sizes: $128, 64, 32$ and $16$. All networks are trained for four warm restart cycles and thus 150 overall epochs after which we have noticed convergence. Other hyper-parameters are a momentum value of $0.9$, a batch-normalization \cite{Ioffe2015} value of $10^{-4}$ to accelerate training and a dropout rate \cite{Srivastava2014} of $0.5$ in the penultimate classification layer. Weights are initialized according to the Kaiming-normal distribution \cite{He2015}. 

We determine a suitable set of hyper-parameters using cross-validation, that is according to the best validation accuracy during the entire training. We then report the test accuracy based on this model. We show the multi-target accuracy's dependency on learning rate and batch size for the two CNN architectures in table \ref{tab:hyperparams_selection}. We notice that the general trend is in favor of lower batch sizes and a learning rate schedule in the range of $\left[10^{-2}, 10^{-5}\right]$. While the evaluated best validation model's test accuracy generally follows a similar trend, the best test accuracies aren't always correlated with a higher validation accuracy, showing a light distribution mismatch between the splits. We further note that the absolute best test accuracy doesn't necessarily coincide with the point of training at which the model achieves the best validation accuracy. In general, the models seem to have a marginally higher accuracy for the test split. The table also shows that validation and test sets are reasonably different from the train set, on which all investigated models achieve an over-fit. 

After determining a suitable set of hyper-parameters, a batch size of 16 and a learning rate cycle between $\left[10^{-2}, 10^{-5}\right]$, we have proceeded with the selection of patch sizes determined through an additional experiment based on best multi-target validation accuracy. We again emphasize that we do not pick hyper-parameters based on test accuracy, even if a model with lower validation accuracy has a better test score.

\subsubsection{Selection of patch size}
\begin{figure}
	\includegraphics[width=\columnwidth]{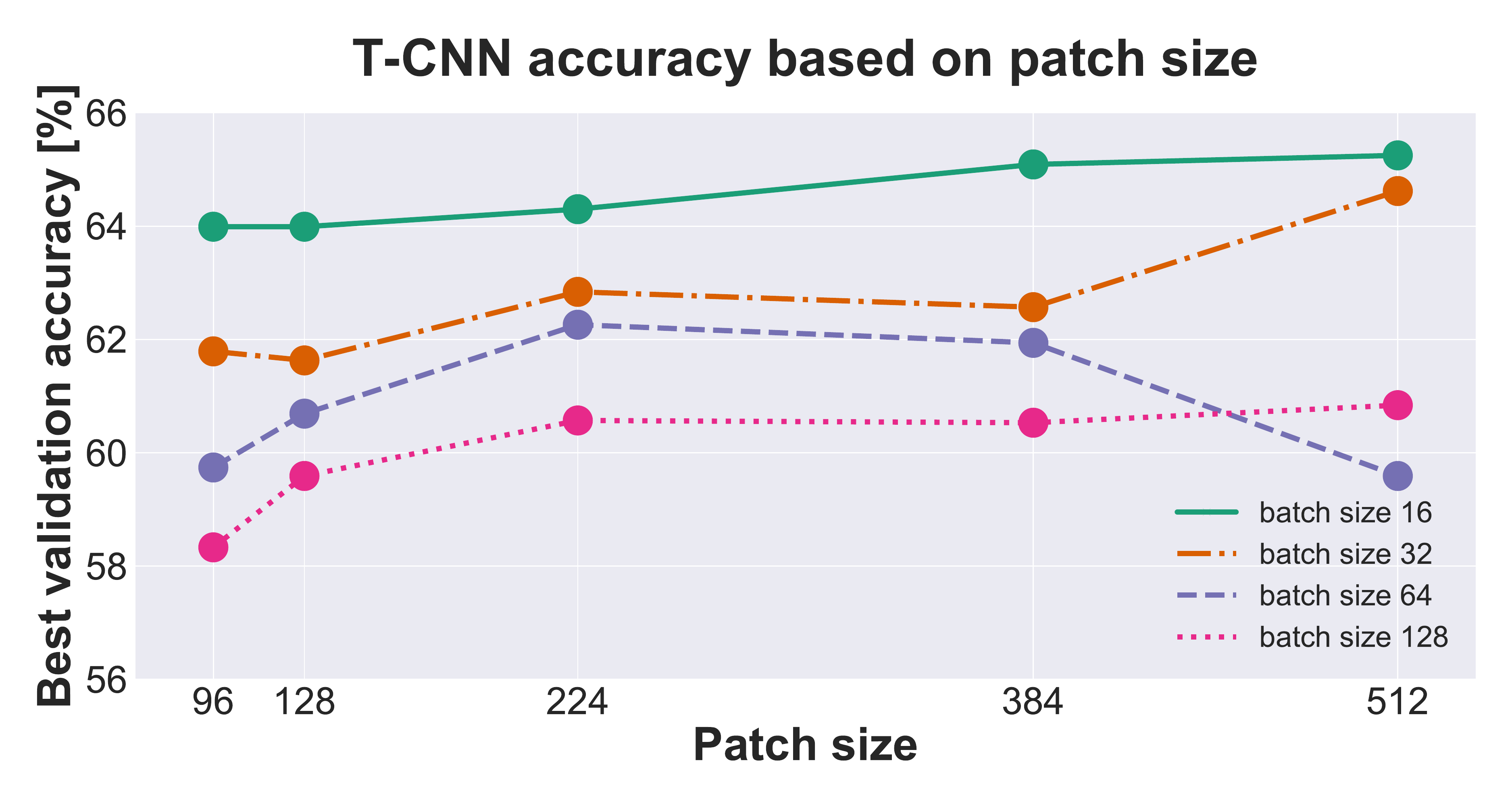} \\
	\includegraphics[width=\columnwidth]{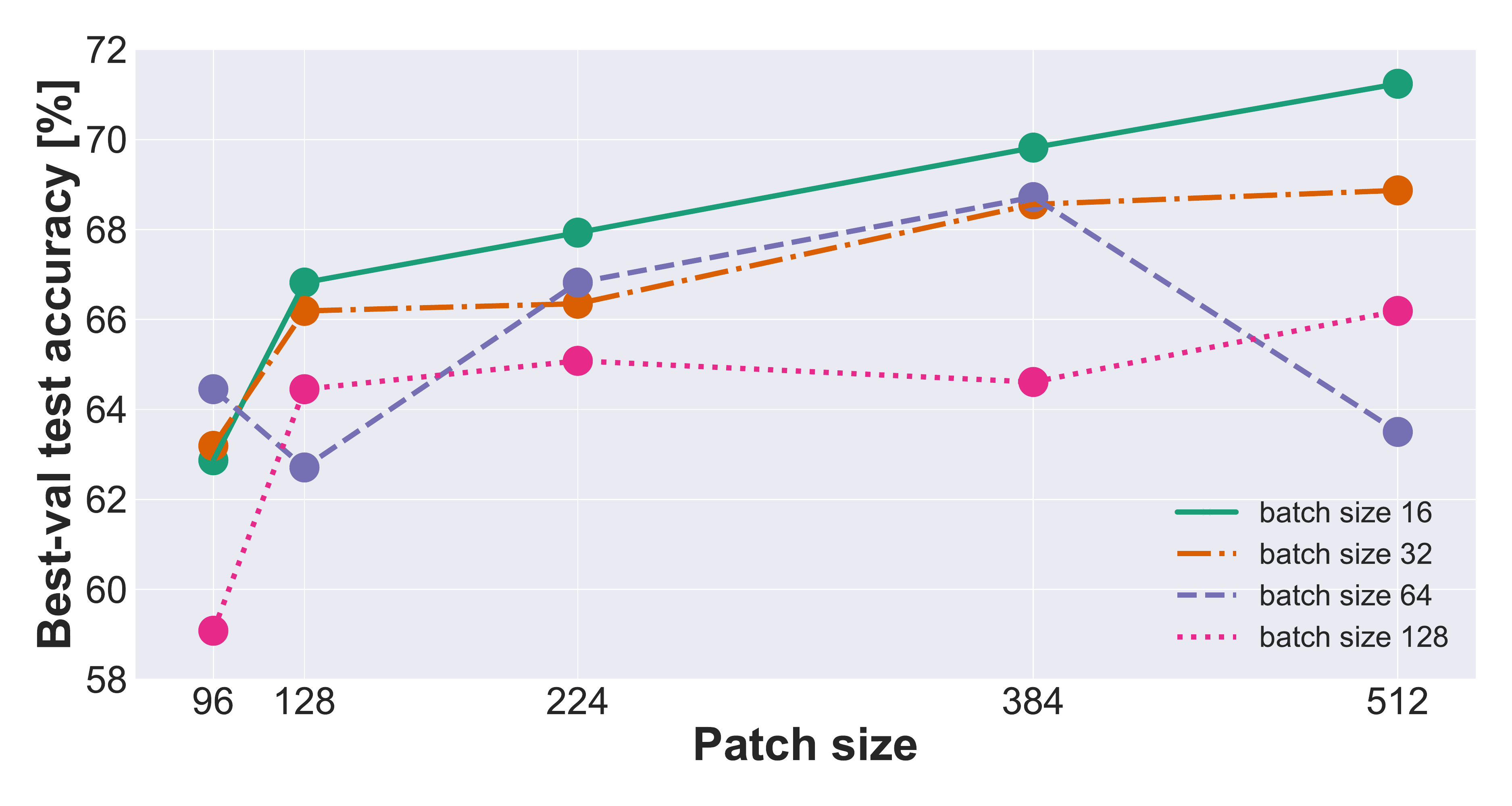}
	\caption{T-CNN multi-target validation accuracy (top panel) and best validation model's multi-target test accuracy (bottom panel) in dependence on patch size.}
	\label{fig:patchsizes}
\end{figure}

Whereas most CNN architectures proposed in the literature are designed for patch sizes of $224 \times 224$, we also evaluate a range of different patch sizes by modifying the number of parameters in the T-CNN model's first fully-connected layer according to the last convolution's spatial output resolution (we do not modify the outgoing feature amounts). In figure \ref{fig:patchsizes} we show the multi-target best validation and corresponding test accuracies for different patch and batch sizes. The perceivable trend is that models trained on patch sizes smaller than $224$ yield less accuracy, whereas the validation accuracy seems to plateau or feature an upwards trend for larger patch sizes. The corresponding test accuracies mirror this trend. We leave the evaluation of even larger patch sizes for future work. For the remainder of this work, we continue to use a patch size of $224$. Although larger patch sizes seem promising they prevent a direct comparison and contrasting of meta-learning approaches with neural network models proposed in the literature without making modifications to their architectures.

\subsubsection{Meta-learning specific parameters}
We design the reward for both MetaQNN and ENAS to fit our multi-target scenario by setting it to the multi-target validation accuracy. We re-iterate that using a per-class accuracy as a metric and particularly to design an RL reward, could lead to controllers being biased towards naively raising the reward by generating models that predict (the easiest) subsets of classes correctly without considering the multi-target overlap properly. We try to set the method specific hyper-parameters of the two meta-learning methods as similar as possible to allow for a direct comparison. We therefore train all child CNN models using the SGDWR schedules and SGD hyper-parameters specified earlier.

\paragraph*{MetaQNN:}
We employ an $\epsilon$-greedy schedule for the Q-learning approach. We train an overall amount of 200 architectures and start with a full exploration phase of $100$ architectures for $\epsilon = 1.0$. We continue with 10 architectures for $\epsilon$ values of $0.9$ to $0.3$ in steps of $0.1$ and finish with 15 architectures for $\epsilon$ values of $0.2$ and $0.1$. Our search space is designed to allow neural architectures with at least $3$ and a maximum of $10$ convolutional layers. We include choices for quadratic filters in the sizes of ${3, 5, 7, 9, 11}$ with possible number of features per layer of ${32, 64, 128, 256}$. We use a Q-learning rate of $0.1$, a discount-factor of $1.0$ and an initial Q-value of $0.15$. The latter is motivated by a $15 \%$ validation accuracy early-stopping criterion at the end of the first SGDWR cycle. In analogy to \cite{Baker2016}, if an architecture doesn't pass this threshold, it is discarded and a new one is sampled and trained. 

Apart from the different reward design, we also make several extensions to the MetaQNN \cite{Baker2016}: We cover down-sampling with an option for convolution stride $s = 2$ for filter sizes larger than $5$. Convolutional layers are further followed by an adaptive pooling stage using spatial-pyramidal pooling (SPP) \cite{He2014} of allowed scales ${3, 4, 5}$ and the possibility to pick a hidden fully-connected layer with size ${32, 64}$ or $128$ before adding the final classification stage. All layers are followed by batch-normalization and a ReLU non-linearity to accelerate training. We also include the possibility to add ResNet-like skip connections between two padded $3 \times 3$ convolutions that do not change spatial dimensionality. If the number of convolutional output features is the same the skip connection is a simple addition, whereas an extra parallel convolution (that isn't counted as an additional layer) is added if the amount of output features needs to change. We make these extensions to provide a fairer comparison to the architecture search of ENAS, that by design contains batch-normalization, adaptive pooling and the possibility of adding skip-connections.

\paragraph*{ENAS:}
In contrast to MetaQNN where the number of layers of each architecture is flexible, network depth in ENAS is pre-determined by the specification of number of nodes in the directed acyclic graph (DAG). Each node defines a possible set of feature operations that the RNN controller samples at each step together with connection patterns. In the process of the search, the same DAG is used to generate architectures with candidates sharing weights through sharing of feature operations. We choose to let the search evolve through alternate training of the CNNs' shared weights on the CODEBRIM train set and the RNN controller's weights on the validation set, where the controller samples one architecture per mini-batch. We design the DAG such that each architecture has $7$ convolutional layers and $1$ classification layer that is followed by a Sigmoid function. We choose this depth to have a direct comparison to the average depth of MetaQNN architectures. The allowed feature operations are convolutions with square filters of size $3$ and $5$, corresponding depth-wise separable convolutions \cite{Chollet2017}, max-pooling and average-pooling with kernel size $3 \times 3$. Each layer is followed by batch-normalization and a ReLU non-linearity. Because ENAS uses shared weights in the search, a final re-training step of proposed architectures is necessary. We use a feature amount of $64$ during the search for all layers and use a DenseNet growth-pattern \cite{Huang2016} of $k = 2$ in the final training consistent with the work of Pham \etal \cite{Pham2018}. The total number of search epochs is $310$ (5 SGDWR cycles) after which we have experienced convergence of the controller. The RNN controller consists of an LSTM \cite{Hochreiter1997} with two hidden-layers of $64$ features that is trained with a learning rate of $10^{-3}$ using ADAM \cite{Kingma2015}. 

\subsection{Results and discussion}

\begin{figure}
	\includegraphics[width=1.05\columnwidth]{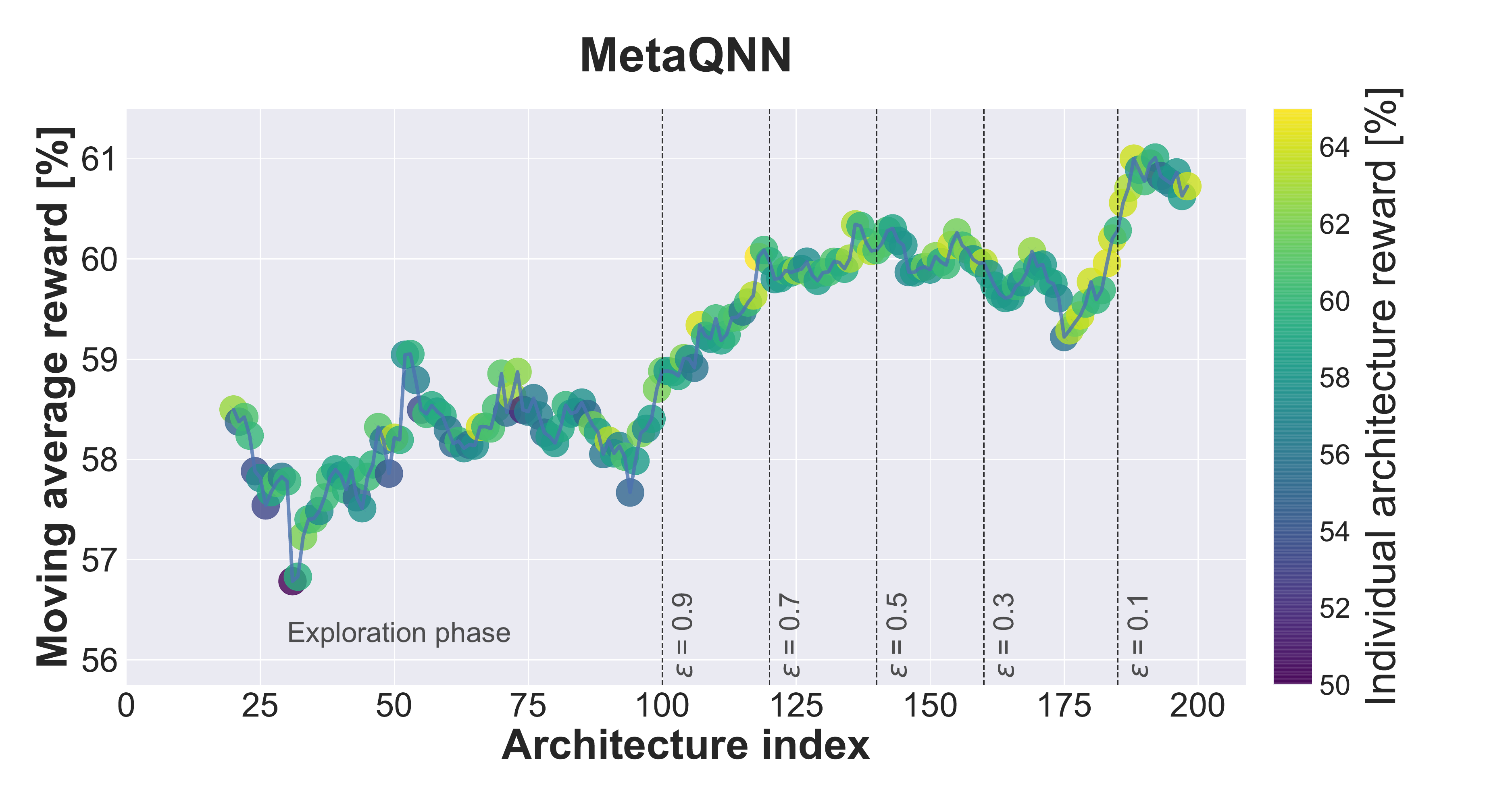} \\
	\includegraphics[width= \columnwidth]{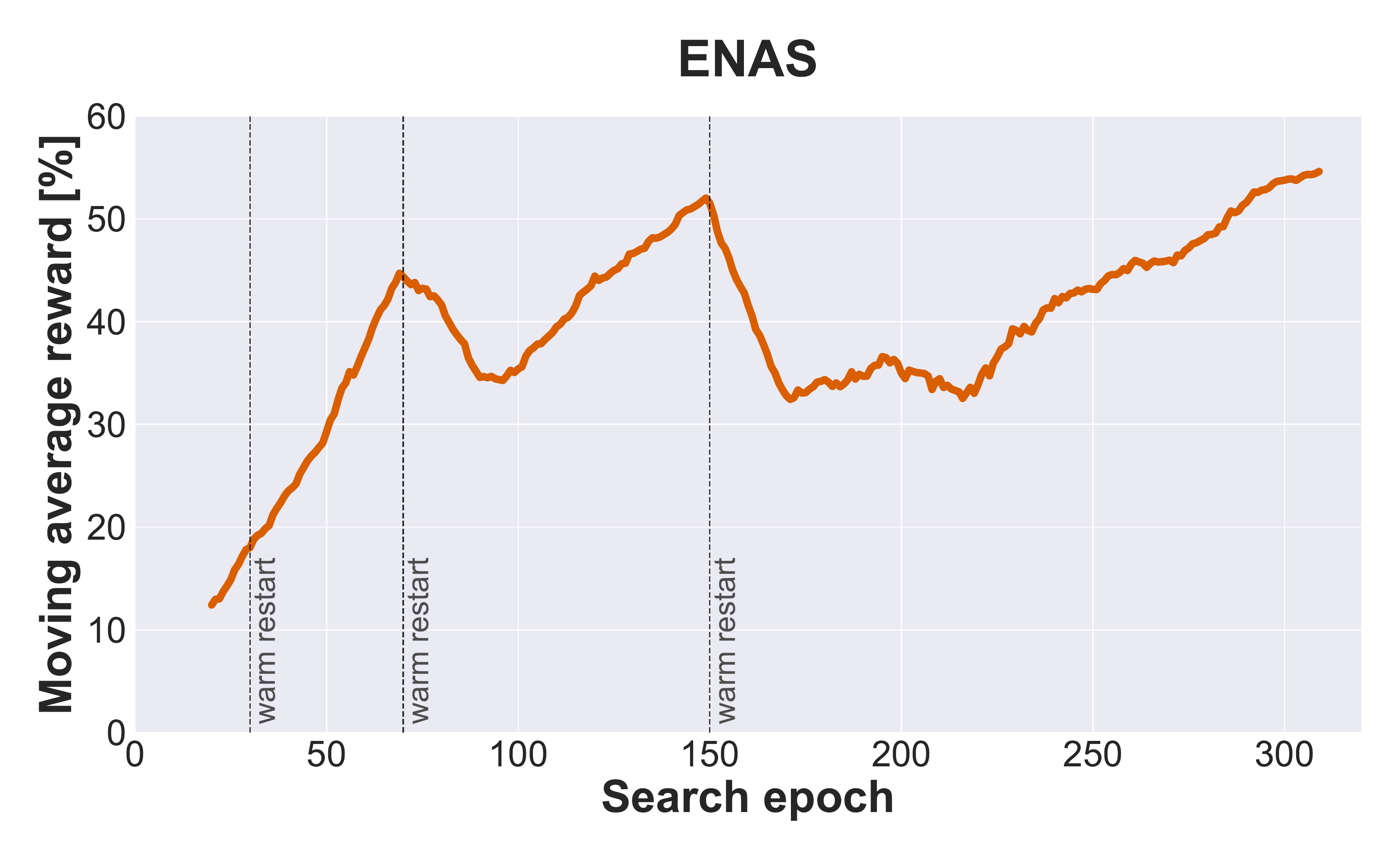}
	\caption{Evolution of the moving average reward defined as the multi-target validation accuracy of architectures proposed through meta-learning. The top panel additionally shows individual architecture accuracies for the MetaQNN in color. ENAS in the bottom panel has shared model weights during training and thus requires a final end-to-end re-training step for final validation accuracies of individual architectures.}
	\label{fig:metalearning}
\end{figure}

\begin{table}
\resizebox{\columnwidth}{!}{
\begin{tabular}{lrrrr}
Architecture & \multicolumn{2}{c}{Multi-target accuracy [\%]} & Params [M] & Layers\\
\cmidrule{2-3}
 & best val & bv-test & & \\ 
\toprule
Alexnet & 63.05 & 66.98 & 57.02 & 8\\ 
T-CNN & 64.30 & 67.93 & 58.60 & 8\\ 
VGG-A & 64.93 & 70.45 & 128.79 & 11\\ 
VGG-D & 64.00 & 70.61 & 134.28 & 16\\ 
WRN-28-4  & 52.51 & 57.19 & 5.84 & 28\\ 
Densenet-121 & 65.56 & 70.77 & 11.50 & 121\\
\midrule
ENAS-1 & 65.47 & 70.78 &  3.41 & 8\\ 
ENAS-2 & 64.53 & 68.91  & 2.71 & 8\\ 
ENAS-3 & 64.38 & 68.75 & 1.70 & 8\\ 
MetaQNN-1 &  66.02 & 68.56  & 4.53 & 6 \\ 
MetaQNN-2 & 65.20 & 67.45  & 1.22 & 8\\ 
MetaQNN-3 & 64.93 & 72.19 & 2.88 & 7\\
\bottomrule
\end{tabular} }
\caption{Comparison of popular CNNs from the literature with the top three architectures of MetaQNN and ENAS in terms of best multi-target validation accuracy (best val), best validation model's test accuracies (bv-test), overall amount of parameters (Params) in million and amount of trainable layers. For WRN we use a width factor of 4 and a growth rate of $k = 32$ for DenseNet.}
\label{tab:performances}
\end{table}

We demonstrate the effectiveness of neural architecture search with MetaQNN and ENAS for multi-target concrete defect classification on the CODEBRIM dataset. We show respective moving average rewards based on a window size of 20 architectures in figure \ref{fig:metalearning}. Individual architecture accuracies for MetaQNN are shown in color for each step in the top panel. We observe that after the initial exploration phase, the Q-learner starts to exploit and architectures improve in multi-target validation accuracy. In the bottom panel of the figure we show corresponding rewards for the shared-weight ENAS DAG. We observe that both methods learn to suggest architectures with improved accuracy over time. We remind the reader that in contrast to the MetaQNN, a final re-training step of the top architectures is needed for ENAS to obtain the task's final accuracy values.

The multi-target validation and test accuracies, again reported at the point in time of best validation, the number of overall architecture parameters and layers for the top three MetaQNN and ENAS architectures can be found in table \ref{tab:performances}. We also evaluate and provide these values for popular CNN baselines: Alexnet \cite{Krizhevsky2012}, VGG \cite{Simonyan2015}, Texture-CNN \cite{Andrearczyk2016}, wide residual networks (WRN) \cite{Zagoruyko2016} and densely connected networks (DenseNet) \cite{Huang2016}.
We see that the Texture-CNN variant of Alexnet slightly outperforms the latter. The connectivity pattern of the DenseNet architecture also boosts the performance in contrast to the VGG models. Lastly, we note that we were only able to achieve heavy over-fitting with WRN configurations (even with other hyper-parameters and other configurations such as WRN-28-10 or WRN-40). 

The accuracies obtained by all of our meta-learned architectures, independently of the underlying algorithm, outperform most baseline CNNs and feature at least similar performance in comparison to DenseNet. Moreover, they feature much fewer parameters with fewer overall layers and are thus more efficient than their computationally heavy counterparts. Our best meta-learned models have validation accuracies as high as $66 \%$, while the test accuracies go up to $72 \%$ with total amount of parameters less than $5$ million. In contrast to literature CNN baselines these architectures are thus more tailored to our specific task and its multi-target nature. Interestingly, previously obtained improvements from one literature CNN baseline to another on ImageNet, such as Alexnet $81.8 \%$ to VGG-D $92.8 \%$ top-5 accuracies, do not show similar improvements when evaluated on our task. This underlines the need for diverse datasets in evaluation of architectural advances and demonstrates how architectures that were hand-designed, even with incredible care and effort, for one dataset such as ImageNet may nonetheless be inferior to meta-learned neural networks.  

Between the two search strategies we do not observe a significant difference in performances. We believe this is due to previously mentioned modifications to MetaQNN, mainly the addition of skip-connections and batch-normalization that make proposed architectures more similar to those of ENAS. We point the reader to the supplementary material for exact definitions of meta-learned architectures. There, we also include a set of image patches that are commonly classified as correct for all targets, images where only part of the overlapping defect classes is predicted and completely misclassified examples. 

\section{Conclusion}
We introduce a novel multi-class multi-target dataset called CODEBRIM for the task of concrete defect recognition. In contrast to previous work that focuses largely on cracks, we classify five commonly occurring and structurally relevant defects through deep learning. Instead of limiting our evaluation to common CNN models from the literature, we adapt and compare two recent meta-learning approaches to identify suitable task-specific neural architectures. Through extension of the MetaQNN, we observe that the two meta-learning techniques yield comparable architectures. We show that these architectures feature fewer parameters, fewer layers and are more accurate than their human designed counterparts on our presented multi-target classification task. Our best meta-learned models achieve multi-target test accuracies as high as $72 \%$. Our work creates prospects for future work such as multi-class multi-target concrete defect detection, semantic segmentation and system applications like UAV based real-time inspection of concrete structures.
\newline
\setlength\intextsep{0pt}
\begin{wrapfigure}[2]{l}{0.055 \textwidth}
	\includegraphics[width=0.055 \textwidth]{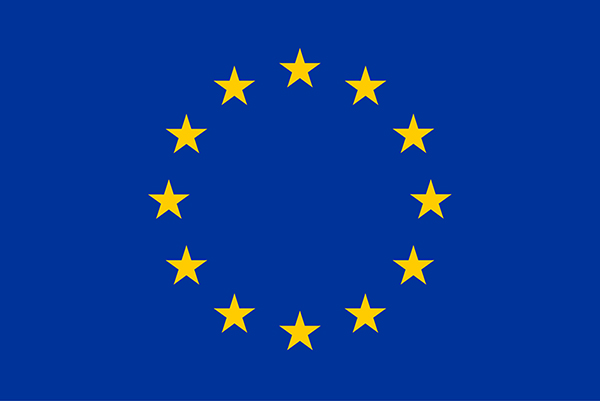}
\end{wrapfigure}
{\small 
\noindent \textbf{Acknowledgements:} This project has received funding from the European Union's Horizon 2020 research and innovation programme under grant agreement No. 687384 "AEROBI". We would like to thank everyone involved in the AEROBI project. Particular appreciation is given for the civil engineering team of Egnatia Odos A.E. and Netivei NTIC, without whom the annotation of the dataset wouldn't have been possible. We further thank FADA-CATEC, Tobias Weis and Sumit Pai for their support in parts of the data acquisition and Hieu Pham for valuable discussion of ENAS hyper-parameters.}

\newpage

{\small
\bibliographystyle{ieee}
\bibliography{references}

\begin{thebibliography}{10}\itemsep=-1pt

\bibitem{Andrearczyk2016}
Vincent Andrearczyk and Paul~F Whelan.
\newblock {Using filter banks in Convolutional Neural Networks for texture
  classification}.
\newblock {\em Pattern Recognition Letters}, 84:63--69, 2016.

\bibitem{Baker2016}
Bowen Baker, Otkrist Gupta, Nikhil Naik, and Ramesh Raskar.
\newblock {Designing Neural Network Architectures using Reinforcement
  Learning}.
\newblock {\em International Conference on Learning Representations (ICLR)},
  2016.

\bibitem{Bell2015}
Sean Bell, Paul Upchurch, Noah Snavely, and Kavita Bala.
\newblock {Material recognition in the wild with the Materials in Context
  Database}.
\newblock In {\em Computer Vision and Pattern Recognition (CVPR)}, 2015.

\bibitem{Cai2018}
Han Cai, Tianyao Chen, Weinan Zhang, Yong Yu, and Jun Wang.
\newblock {Efficient Architecture Search by Network Transformation}.
\newblock {\em AAAI Conference on Artificial Intelligence (AAAI)}, 2018.

\bibitem{Caputo2005}
Barbara Caputo, Eric Hayman, and P Mallikarjuna.
\newblock {Class-specific material categorisation}.
\newblock In {\em International Conference on Computer Vision (ICCV)}, 2005.

\bibitem{Chollet2017}
Francois Chollet.
\newblock {Xception: Deep Learning with Depthwise Seperable Convolutions}.
\newblock In {\em Computer Vision and Pattern Recognition (CVPR)}, pages
  1800--1807, 2017.

\bibitem{Cimpoi2015}
Mircea Cimpoi, Subhransu Maji, and Andrea Vedaldi.
\newblock {Deep convolutional filter banks for texture recognition and
  segmentation}.
\newblock In {\em Computer Vision and Pattern Recognition (CVPR)}, 2015.

\bibitem{DaSilva2018}
Wilson R.~L. da Silva and Diogo~S. de Lucena.
\newblock {Concrete Cracks Detection Based on Deep Learning Image
  Classification}.
\newblock In {\em International Conference on Experimental Mechanics (ICEM18)},
  2018.

\bibitem{Dana1999}
Kristin~J. Dana, Bram van Ginneken, Shree~K. Nayar, and Jan~J. Koenderink.
\newblock {Reflectance and texture of real-world surfaces}.
\newblock {\em ACM Transactions on Graphics (TOG)}, 18(1):1--34, 1999.

\bibitem{Everingham2015}
Mark Everingham, S.~M.Ali~Ali Eslami, Luc {Van Gool}, Christopher K.I.~I
  Williams, John Winn, and Andrew Zisserman.
\newblock {The Pascal Visual Object Classes Challenge: A Retrospective}.
\newblock {\em International Journal of Computer Vision (IJCV)},
  111(1):98--136, 2014.

\bibitem{Hayman2004}
Eric Hayman, Barbara Caputo, Mario Fritz, and Jan-Olof Eklundh.
\newblock {On the Significance of Real-World Conditions for Material
  Classification}.
\newblock In {\em European Conference on Computer Vision (ECCV)}, 2004.

\bibitem{He2014}
Kaiming He, Xiangyu Zhang, Shaoqing Ren, and Jian Sun.
\newblock {Spatial Pyramid Pooling in Deep Convolutional Networks for Visual
  Recognition}.
\newblock In {\em European Conference on Computer Vision (ECCV)}, pages
  346--361, 2014.

\bibitem{He2015}
K. He, X. Zhang, S. Ren, and J. Sun.
\newblock {Delving deep into rectifiers: Surpassing human-level performance on
  imagenet classification}.
\newblock In {\em International Conference on Computer Vision (ICCV)}, pages
  1026--1034, 2015.

\bibitem{He2016}
K. He, X. Zhang, S. Ren, and J. Sun.
\newblock {Deep Residual Learning for Image Recognition}.
\newblock In {\em Computer Vision and Pattern Recognition (CVPR)}, 2016.

\bibitem{Hochreiter1997}
Sepp Hochreiter and J{\"{u}}rgen Schmidhuber.
\newblock {Long Short-Term Memory}.
\newblock {\em Neural Computation}, 9(8):1735--1780, 1997.

\bibitem{Huang2016}
Gao Huang, Zhuang Liu, Laurens {Van Der Maaten}, and Kilian~Q. Weinberger.
\newblock {Densely connected convolutional networks}.
\newblock In {\em Computer Vision and Pattern Recognition (CVPR)}, pages
  2261--2269, 2017.

\bibitem{Ioffe2015}
S. Ioffe and C. Szegedy.
\newblock {Batch Normalization: Accelerating Deep Network Training by Reducing
  Internal Covariate Shift}.
\newblock In {\em International Conference on Machine Learning (ICML)},
  volume~37, pages 448--456, 2015.

\bibitem{Kim2018}
Hyunjun Kim, Eunjong Ahn, Myoungsu Shin, and Sung-Han Sim.
\newblock {Crack and Noncrack Classification from Concrete Surface Images Using
  Machine Learning}.
\newblock {\em Structural Health Monitoring}, 2018.

\bibitem{Kingma2015}
Diederik~P. Kingma and Jimmy~Lei Ba.
\newblock {Adam: a Method for Stochastic Optimization}.
\newblock In {\em International Conference on Learning Representations (ICLR)},
  2015.

\bibitem{Koch2015}
Christian Koch, Kristina Georgieva, Varun Kasireddy, Burcu Akinci, and Paul
  Fieguth.
\newblock {A review on computer vision based defect detection and condition
  assessment of concrete and asphalt civil infrastructure}.
\newblock {\em Advanced Engineering Informatics}, 29(2):196--210, 2015.

\bibitem{Krizhevsky2012}
Alex Krizhevsky, Ilya Sutskever, and Geoffrey~E. Hinton.
\newblock {ImageNet Classification with Deep Convolutional Neural Networks}.
\newblock In {\em Neural Information Processing Systems (NeurIPS)}, volume~25,
  pages 1097--1105, 2012.

\bibitem{Lecun2015}
Yann Lecun, Yoshua Bengio, and Geoffrey Hinton.
\newblock {Deep learning}.
\newblock {\em Nature}, 521(7553):436--444, 2015.

\bibitem{Li2018}
Yundong Li, Hongguang Li, and Hongren Wang.
\newblock {Pixel-wise crack detection using deep local pattern predictor for
  robot application}.
\newblock {\em Sensors}, 18(9), 2018.

\bibitem{Liu2018}
Hanxiao Liu, Karen Simonyan, and Yiming Yang.
\newblock {DARTS: Differentiable Architecture Search}.
\newblock {\em International Conference on Learning Representations (ICLR)},
  2019.

\bibitem{Loshchilov2017}
I. Loshchilov and F. Hutter.
\newblock {SGDR: Stochastic Gradient Descent With Warm Restarts}.
\newblock In {\em International Conference on Learning Representations (ICLR)},
  2017.

\bibitem{Macquire2018}
Marc Macquire, Sattar Dorafshan, and Robert~J. Thomas.
\newblock {SDNET2018: A concrete crack image dataset for machine learning
  applications}.
\newblock {\em https://digitalcommons.usu.edu/all{\_}datasets/48 (last access:
  06.11.18)}, Paper 48, 2018.

\bibitem{Pham2018}
Hieu Pham, Melody~Y. Guan, Barret Zoph, Quoc~V. Le, and Jeff Dean.
\newblock {Efficient Neural Architecture Search via Parameters Sharing}.
\newblock In {\em International Conference on Machine Learning (ICML)}, 2018.

\bibitem{Real2017}
Esteban Real, Sherry Moore, Andrew Selle, Saurabh Saxena, Yutaka~Leon Suematsu,
  Quoc Le, and Alex Kurakin.
\newblock {Large-Scale Evolution of Image Classifiers}.
\newblock In {\em International Conference on Machine Learning (ICML)}, 2017.

\bibitem{ILSVRC}
Olga Russakovsky, Jia Deng, Hao Su, Jonathan Krause, Sanjeev Satheesh, Sean Ma,
  Zhiheng Huang, Andrej Karpathy, Aditya Khosla, Michael Bernstein,
  Alexander~C. Berg, and Li Fei-Fei.
\newblock {ImageNet Large Scale Visual Recognition Challenge}.
\newblock In {\em International Journal of Computer Vision (IJCV)}, volume 115,
  pages 211--252, 2015.

\bibitem{Sharan2009}
Lavanya Sharan, Ruth Rosenholtz, and Edward~H. Adelson.
\newblock {Material perception: What can you see in a brief glance?}
\newblock {\em Journal of Vision (JOV)}, 9(8), 2009.

\bibitem{Shi2016}
Yong Shi, Limeng Cui, Zhiquan Qi, Fan Meng, and Zhensong Chen.
\newblock {Automatic road crack detection using random structured forests}.
\newblock {\em IEEE Transactions on Intelligent Transportation Systems
  (T-ITS)}, 17(12):3434--3445, 2016.

\bibitem{Simonyan2015}
K. Simonyan and A. Zisserman.
\newblock {Very Deep Convolutional Networks for Large-Scale Image Recognition}.
\newblock In {\em International Conference on Learning Representations (ICLR)},
  2015.

\bibitem{Srivastava2014}
Nitish Srivastava, Geoffrey~E. Hinton, Alex Krizhevsky, Ilya Sutskever, and
  Ruslan Salakhutdinov.
\newblock {Dropout : A Simple Way to Prevent Neural Networks from Overfitting}.
\newblock {\em Journal of Machine Learning Research (JMRL)}, 15:1929--1958,
  2014.

\bibitem{Sutton1999}
S.~Richard Sutton, David McAllester, Satinder Singh, and Yishay Mansour.
\newblock {Policy Gradient Methods for Reinforcement Learning with Function
  Approximation}.
\newblock In {\em Neural Information Processing Systems (NeurIPS)}, pages
  1057--1063, 1999.

\bibitem{Tzutalin2015}
Tzutalin.
\newblock {LabelImg}.
\newblock {\em https://github.com/tzutalin/labelImg}, 2015.

\bibitem{Xiao2010}
Jianxiong Xiao, James Hays, Krista~A Ehinger, and Antonio Torralba.
\newblock {SUN Database : Large-scale Scene Recognition from Abbey to Zoo}.
\newblock In {\em Computer Vision and Pattern Recognition (CVPR)}, 2010.

\bibitem{Yang2017}
Liang Yang, Bing Li, Wei Li, Zhaoming Liu, Guoyong Yang, and Jizhong Xiao.
\newblock {Deep Concrete Inspection Using Unmanned Aerial Vehicle Towards CSSC
  Database}.
\newblock In {\em International Conference on Intelligent Robots and Systems
  (IROS)}, 2017.

\bibitem{Zagoruyko2016}
Sergey Zagoruyko and Nikos Komodakis.
\newblock {Wide Residual Networks}.
\newblock In {\em British Machine Vision Conference (BMVC)}, pages 87.1--87.12,
  2016.

\bibitem{Zhang2017}
Yan Zhang, Mete Ozay, Xing Liu, and Takayuki Okatani.
\newblock {Integrating deep features for material recognition}.
\newblock In {\em International Conference on Pattern Recognition (ICPR)},
  2016.

\bibitem{Zoph2017}
Barret Zoph and Quoc~V. Le.
\newblock {Neural Architecture Search with Reinforcement Learning}.
\newblock In {\em International Conference on Learning Representations (ICLR)},
  2017.

\end{thebibliography}
}

\clearpage
\normalfont

\appendix
\section{Content overview}
\noindent \blfootnote{* work conducted while at Frankfurt Institute for Advanced Studies} The supplementary material contains further details for material presented in the main body. 

We start with an extended description of the CODEBRIM dataset in section \ref{supp:dataset}. Here, we provide the specific settings for the employed cameras for dataset image acquisition. In addition to the histogram presented in the main body that shows number of different defects per bounding box, we further provide a histogram with amount of bounding box annotations per image. Additional material reveals specifics of the main body's figure depicting the large variations in distribution of bounding boxes by illustrating the individual nuances of this distribution per defect class. The supplementary dataset material is concluded with a brief discussion on background patch generation.

In supplementary section \ref{supp:dl} we provide a brief discussion on why multi-target accuracy is a better reward metric than naive single-class accuracies and show what multi-target accuracies would translate to in terms of a naive average single-class accuracy. We give detailed descriptions and graphs of the six meta-learned architectures for the top three models obtained through MetaQNN and ENAS. Although it isn't an immediate extension to the main body, but rather additional content, we provide a compact section on transfer learning with experiments conducted with models pre-trained on the ImageNet and MINC datasets. We have decided to move these experiments to the the supplementary material for the interested reader as they do not show any improvements over the content presented in the main body. We conclude the supplementary material with examples for images that are commonly classified correctly as well as showing some typical false multi-target classifications to give the reader a better qualitative understanding of the dataset complexity and challenges. 

\section{CODEBRIM dataset}\label{supp:dataset}

\subsection{Delamination as a defect class}
Some of the CODEBRIM dataset features images that have a defect that is typically referred to as delimation. It is a stage where areas start to detach from the surface. Delamination can thus be recognized by a depth offset of a layer from the main surface body. However, in images acquired by a single camera, especially if the images were acquired using a camera view direction that is orthogonal to the surface, these boundaries are often visually not distinguishable from cracks. Without further information, even a civil engineering expert faces major difficulty in such a distinction between these categories. We have thus decided to label eventual occurrences of delamination together with the crack category. 

\subsection{Cameras}
\begin{table*}[t]
\centering
\resizebox{\textwidth}{!}{\begin{tabular}{lrrrrrrrr}
\textbf{Camera} & \textbf{Resolution [pixels]} & \textbf{Exposure [s]} & \textbf{f [mm]} & \textbf{F-value [f/]} & \textbf{ISO} & \textbf{Flash} \\
\toprule  
Canon IXUS 870 IS & $2592 \times 1944$ & flexible & $5 - 20$ & $2.8 - 5.8$ & $100 - 800$ & none \\ 
Panasonic DMC-FZ72 & $4608 \times 3456$ & $1/250$ & $4 - 42$ & $5.6$ & $400$ & built-in\\ 
Nikon  D5200 & $6000 \times 4000$ & $1/200$ & $55$ & $11.0$ & $200$ & built-in\\ 
Sony $\alpha$-6000 & $6000 \times 3376$ & $1/1000$ & $50$ & $2.0-5.6$ & $50-400$ & HVL-F43M \\
\bottomrule
\end{tabular}}
\caption{Description of cameras, including
resolution, exposure time in fraction of a second, focal length $f$ in mm, the aperture or F-value in terms of focal length,
ISO speed rating and information on potentially used flash.}
\label{tab:cameras}
\end{table*}
We show the four cameras used for acquisition of dataset images in table \ref{tab:cameras}. All chosen cameras have a resolution above Full-HD, with the highest resolution going up to $6000 \times 4000$ pixels. For two cameras we have used a lens with varying focal length, whereas two cameras had a lens with fixed focal length of $50$ and $\SI{55}{\milli\metre} $ respectively.  We have further systematically varied aperture in conjunction with the use of diffused flash modules to homogeneously illuminate dark bridge areas, while also adjusting for changing global illumination (avoiding heavy over or under-exposure). A different very crucial aspect was the employed exposure time. Pictures acquired by UAV were generally captured with a much shorter exposure time to avoid blurring of the image due to out of focus acquisition or inherent vibration and movement of the UAV. One of our cameras, Sony $\alpha$-6000 has thus exclusively been used in the context of UAV based image acquisition with an exposure time of $1/1000$ seconds.  

\begin{figure}[t]
	\centering
	\includegraphics[width=\columnwidth]{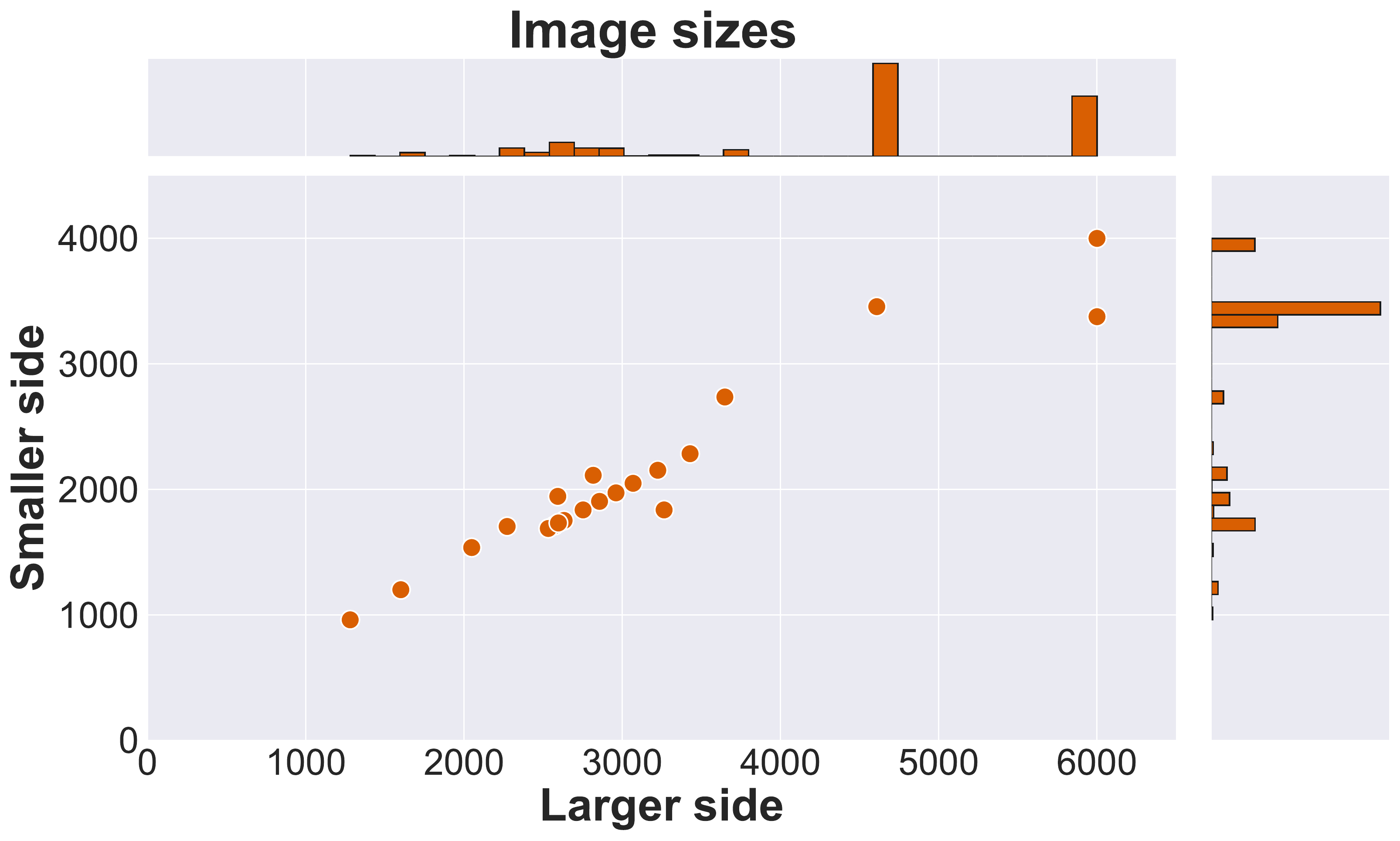} 
	\caption{Distribution of image resolutions. Smaller and larger side refer to the image's larger and smaller dimension.}
	\label{fig:image_sizes}
\end{figure}
We show how the CODEBRIM dataset is practically comprised of the varying resolutions resulting from use with different cameras and settings in figure \ref{fig:image_sizes}. We can observe that the aspect ratio is almost constant with changes in absolute resolution and that the large majority of images has been acquired at very high resolutions. 

\subsection{Annotation process}
\begin{figure}[t]
	\includegraphics[width= \columnwidth]{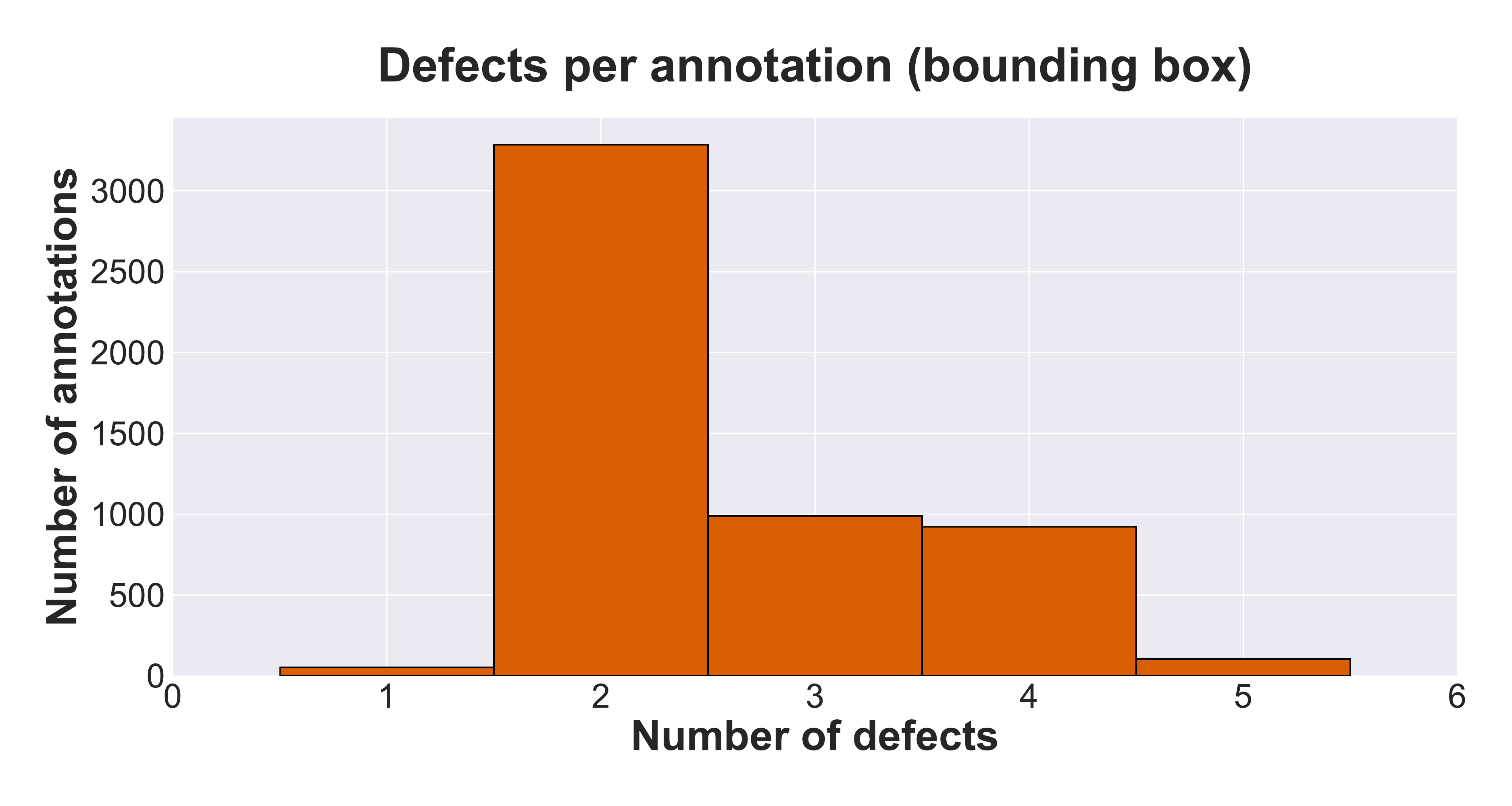}
	\caption{Histogram of number of simultaneously occurring defect classes per annotated bounding box.}
	\label{fig:annotations}
\end{figure}

After curating acquired images by excluding the majority of images that do not have defects, we employed a multi-stage annotation process to create a multi-class multi-target classification dataset using the annotation tool LabelImg  \cite{Tzutalin2015} in consultation with civil engineering experts:

\begin{enumerate}
	\item We first annotated bounding boxes for areas containing defects in the Pascal format \cite{Everingham2015}. 
	\item Each individual bounding box was analyzed with respect to one defect class and a corresponding label was set if the defect is present. 
	\item After finishing the entire set of bounding boxes for one class, we repeated step 2 for the remaining classes and arrived at a multi-class multi-target annotation.
	\item In the last stage, we sampled bounding boxes containing background (concrete without defects as well as non-concrete) according to absolute count, aspect ratios and size of annotated defect bounding boxes.
\end{enumerate}
The reason for staging the process is that we found the annotation process to be less error prone if annotators had to concentrate on the presence of one defect at a time. 

\subsection{Further dataset statistics}
\begin{figure}[t]
	\centering
	\includegraphics[width= \columnwidth]{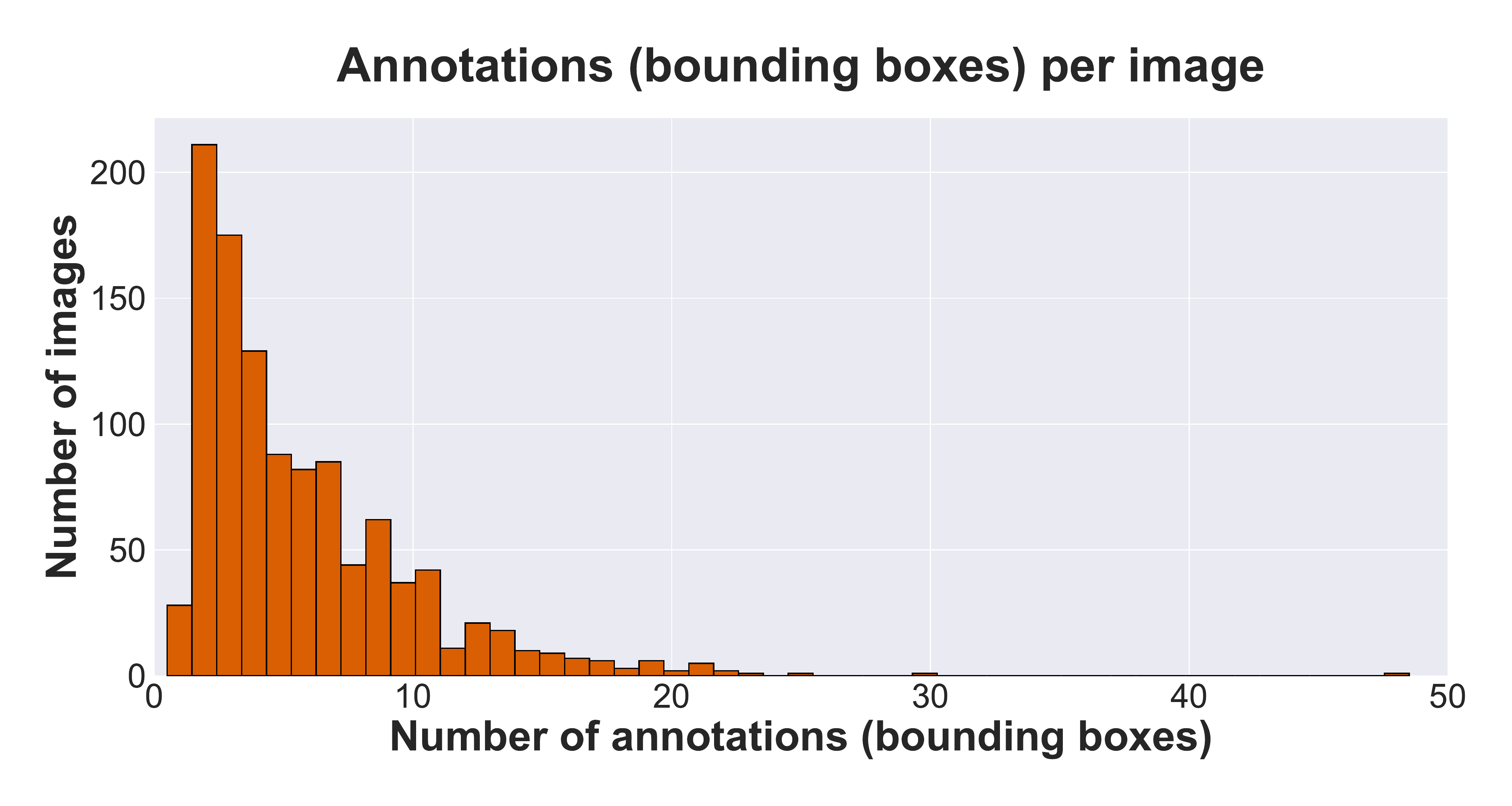} 
	\caption{Distribution of number of bounding box annotations per image.}
	\label{fig:annotations_per_image}
\end{figure}
\begin{figure*}[t]
	\centering
	\includegraphics[width=0.33 \textwidth]{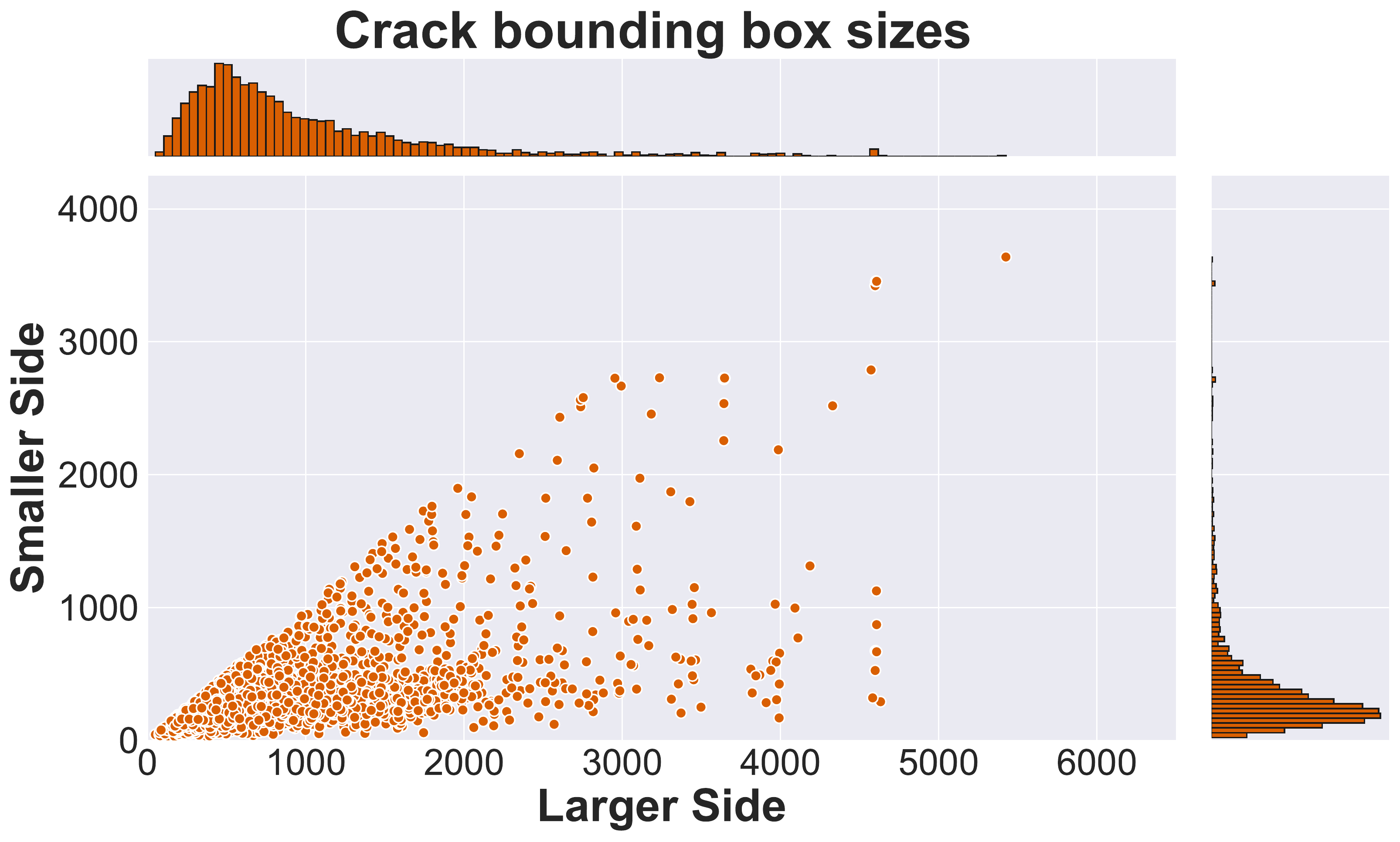}
	\includegraphics[width=0.33 \textwidth]{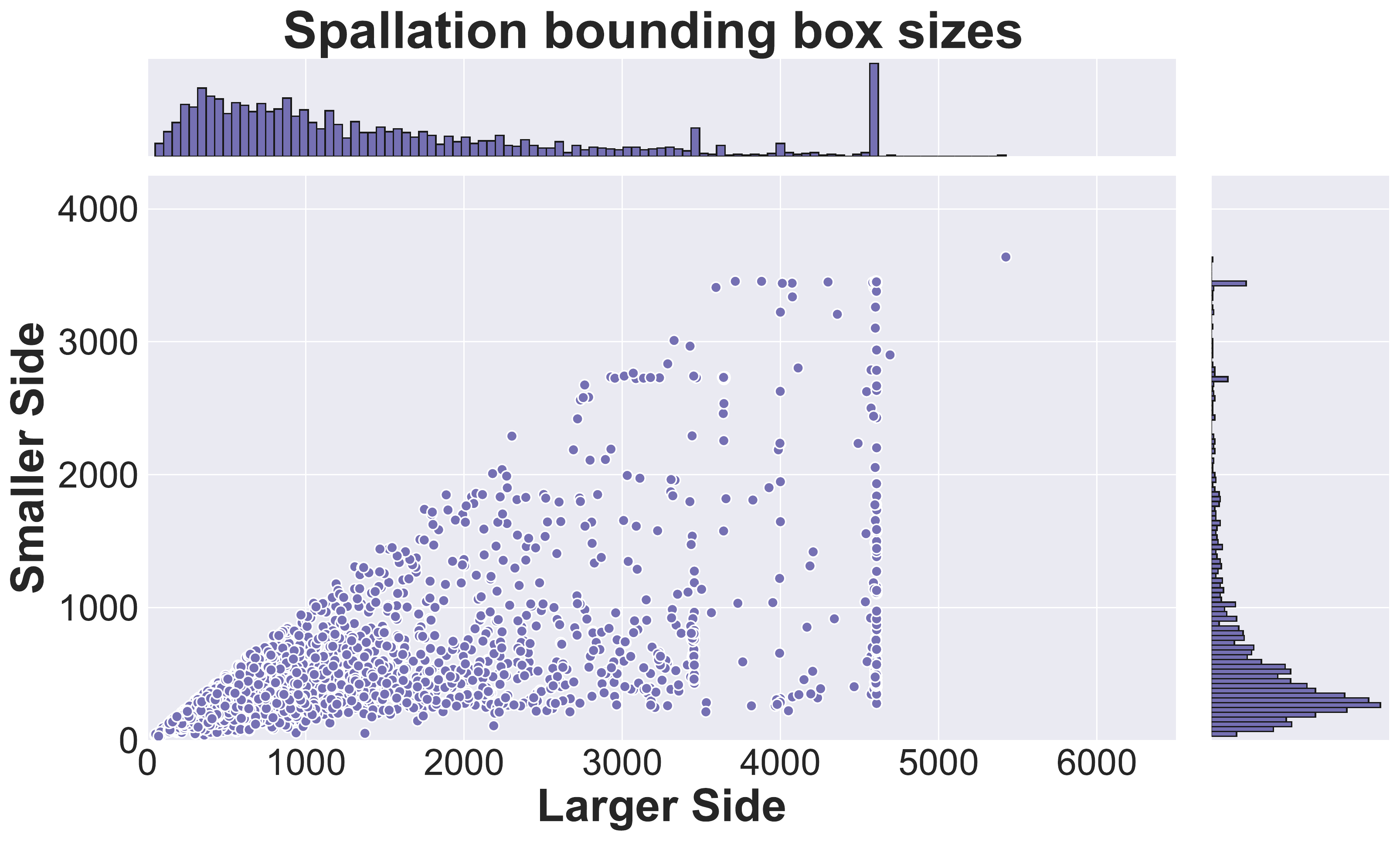}
	\includegraphics[width=0.33 \textwidth]{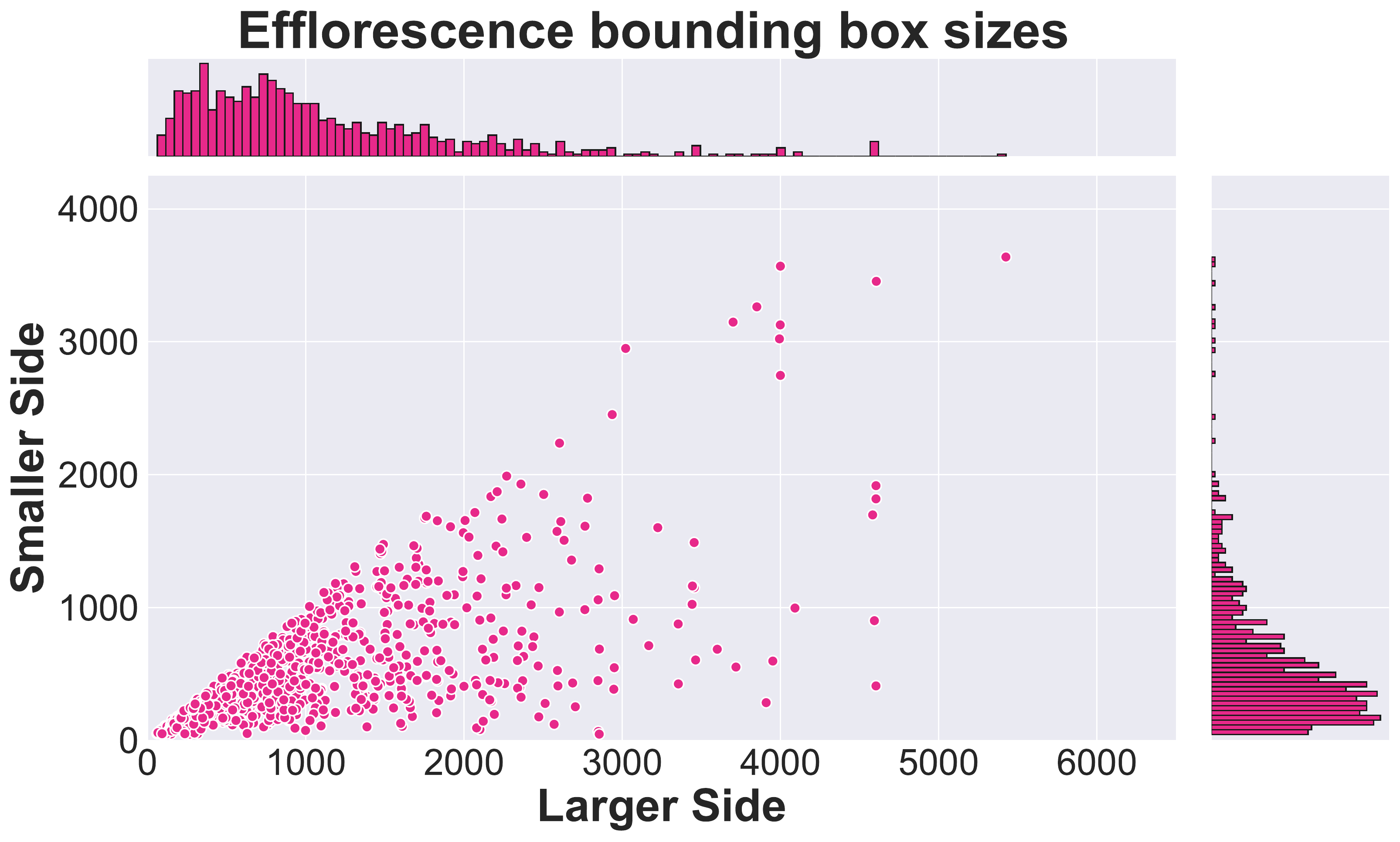} \\
	\includegraphics[width=0.33 \textwidth]{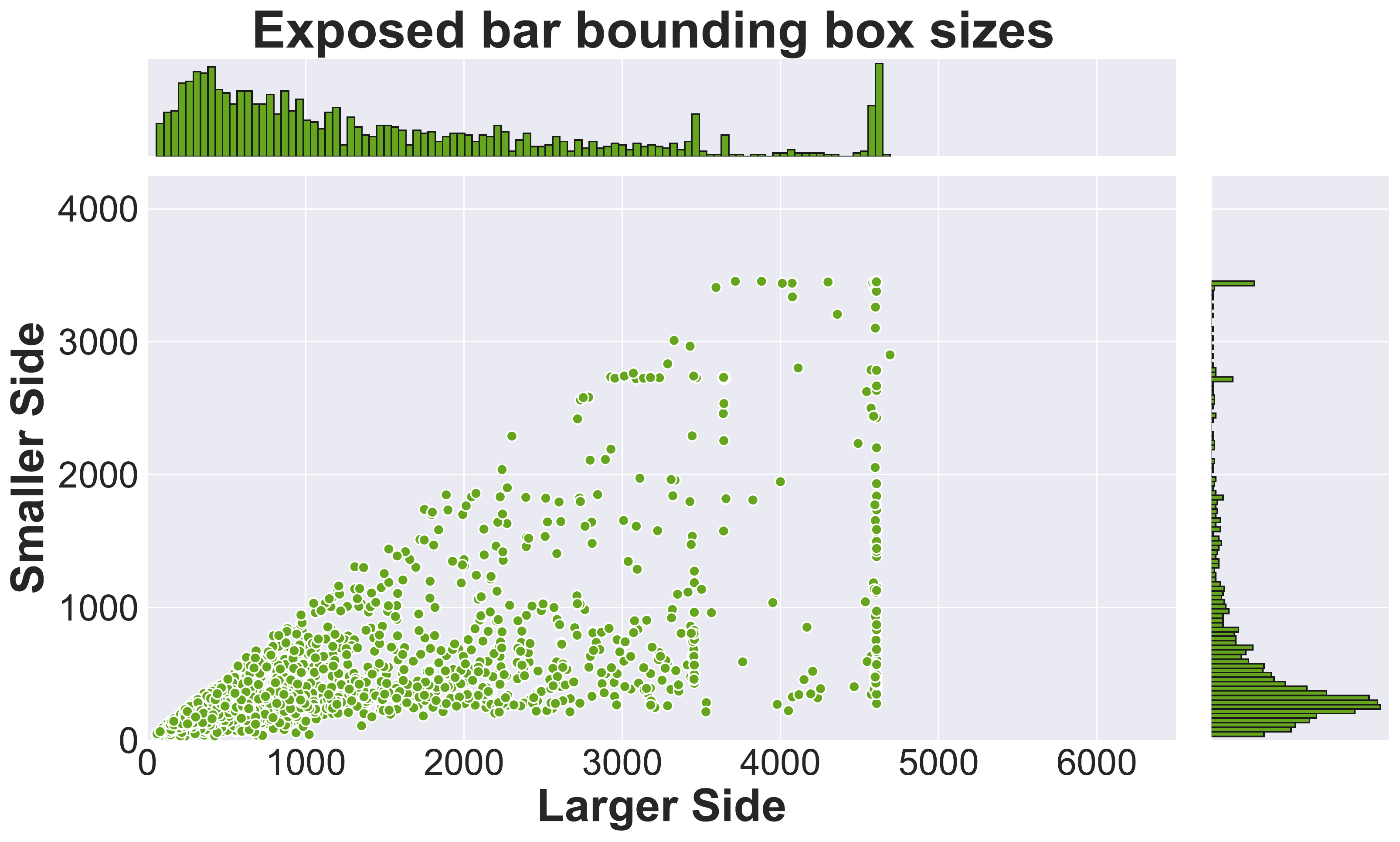}
	\includegraphics[width=0.33 \textwidth]{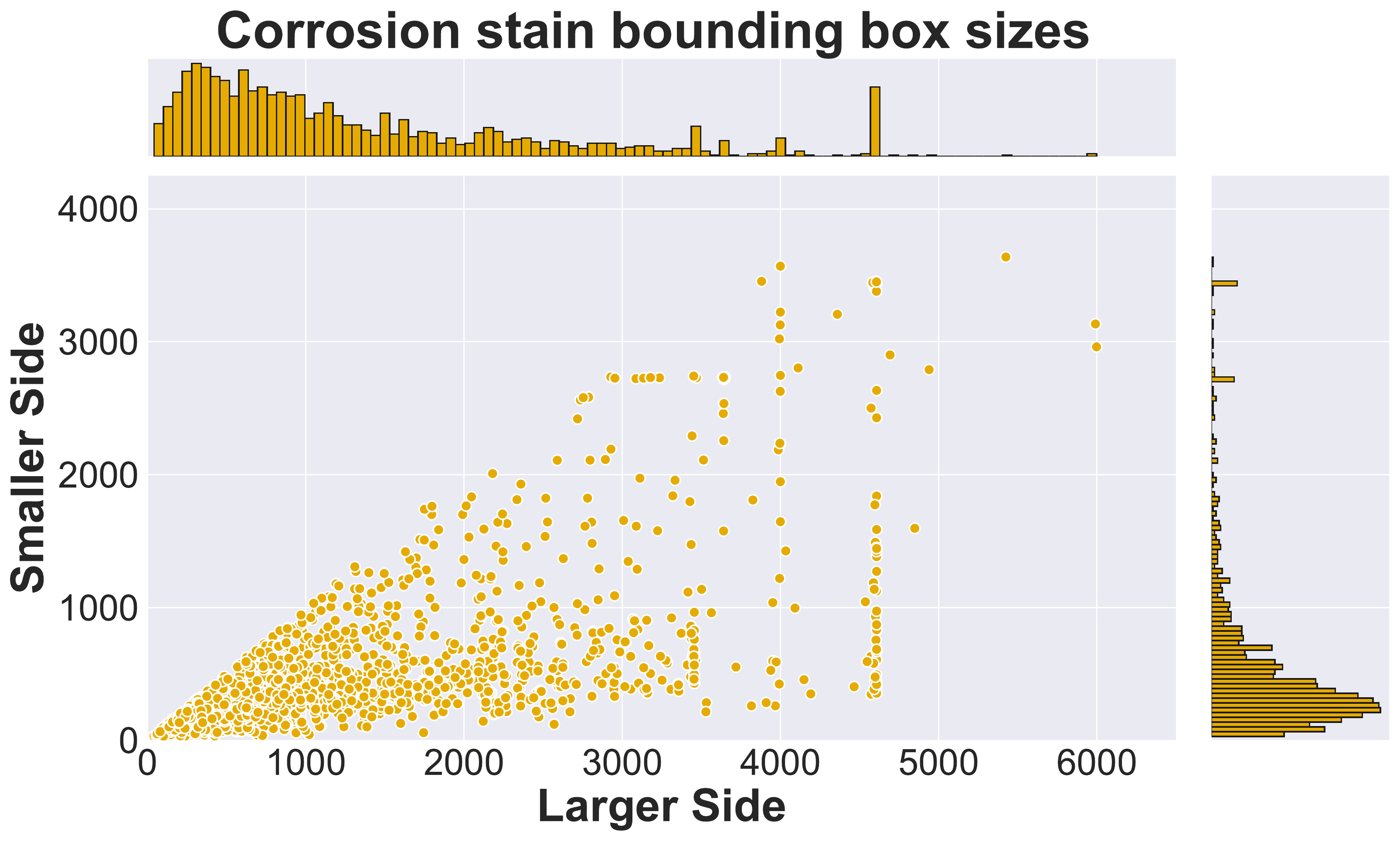}
	
	\caption{Individual distributions of annotated bounding box
sizes for each of the 5 defect classes.}
	\label{fig:sizes_per_defect}
\end{figure*}

\begin{figure*}[t]
	\centering
	\includegraphics[width=0.33 \textwidth]{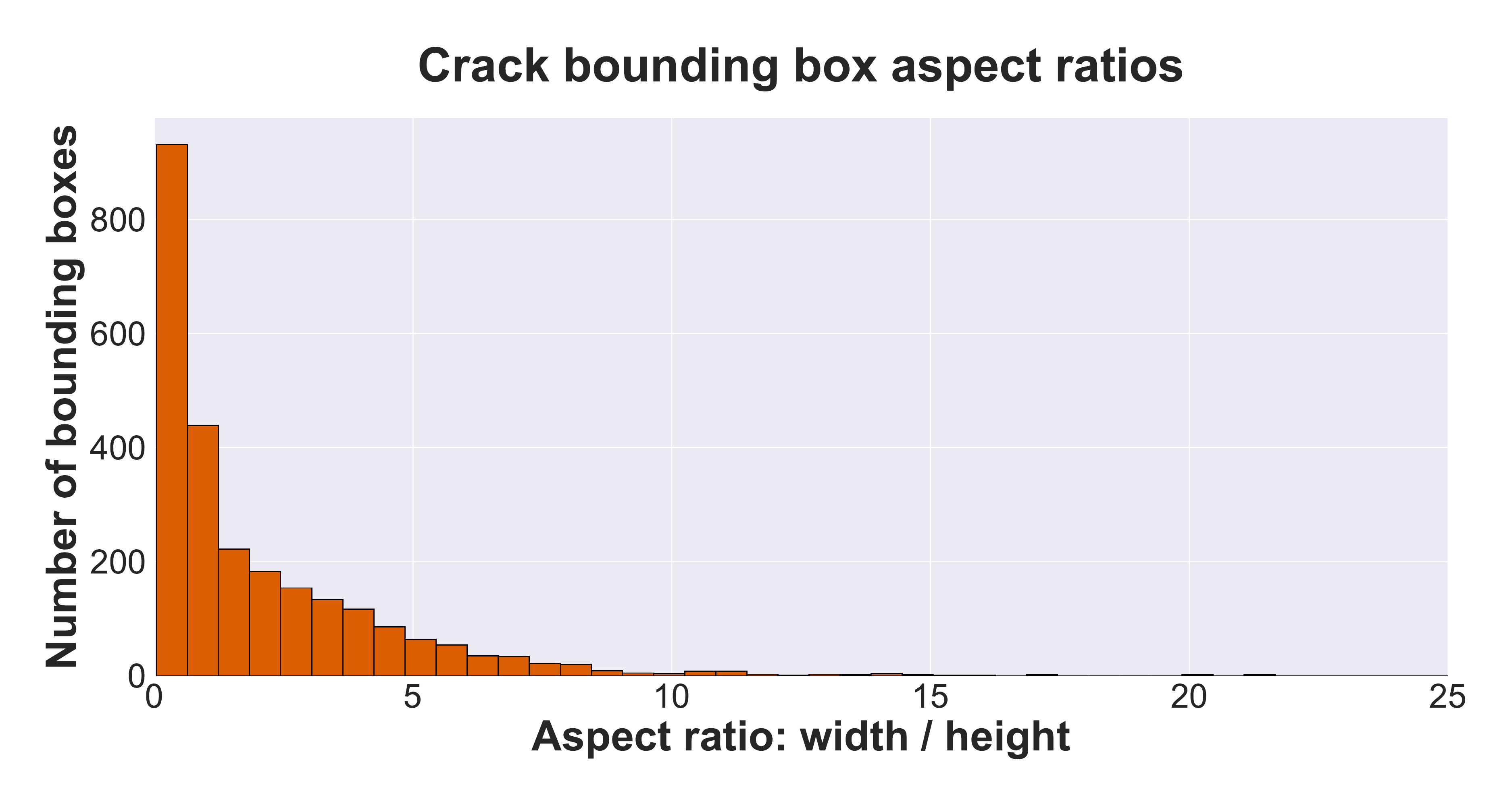}
	\includegraphics[width=0.33 \textwidth]{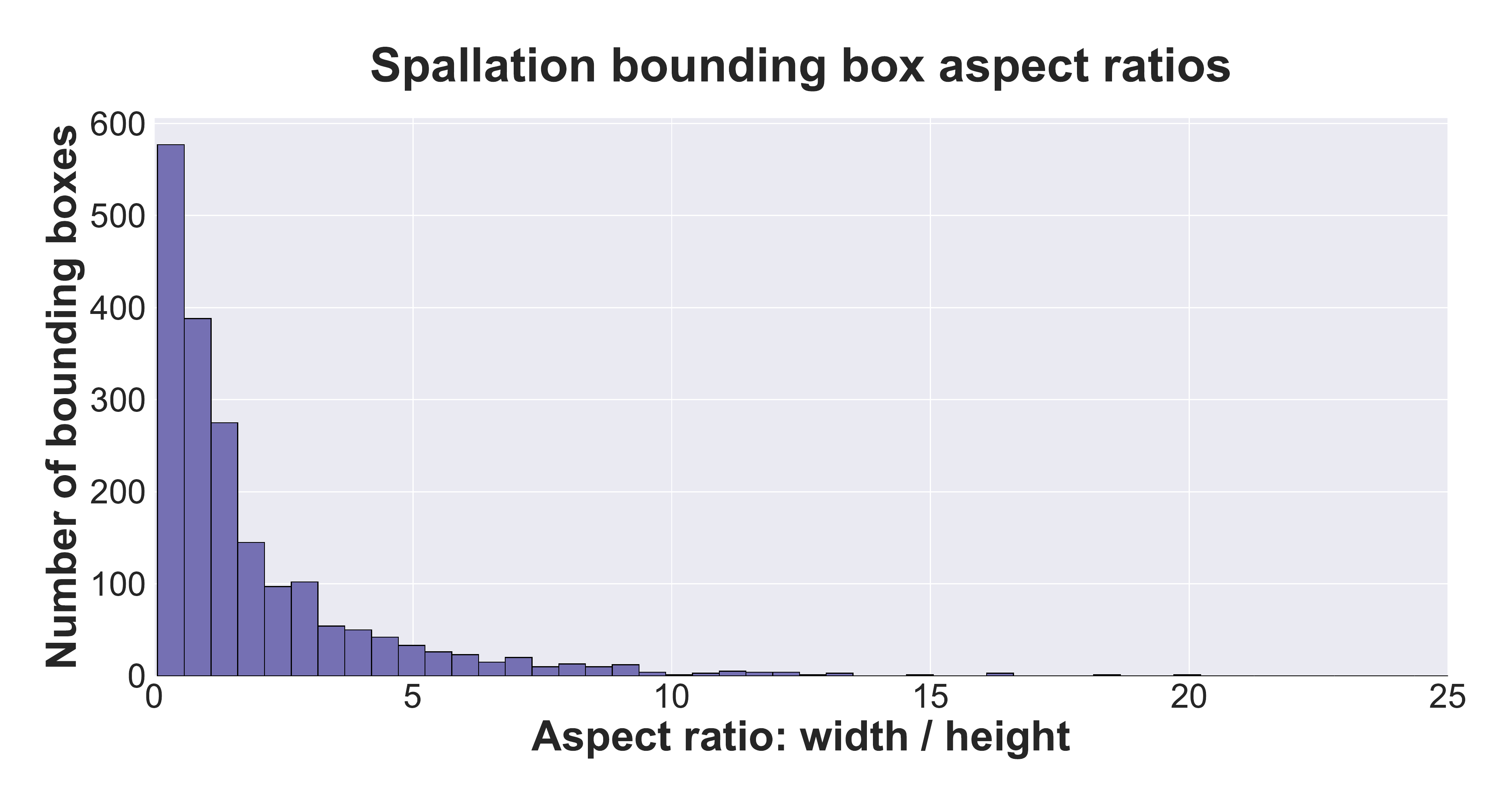}
	\includegraphics[width=0.33 \textwidth]{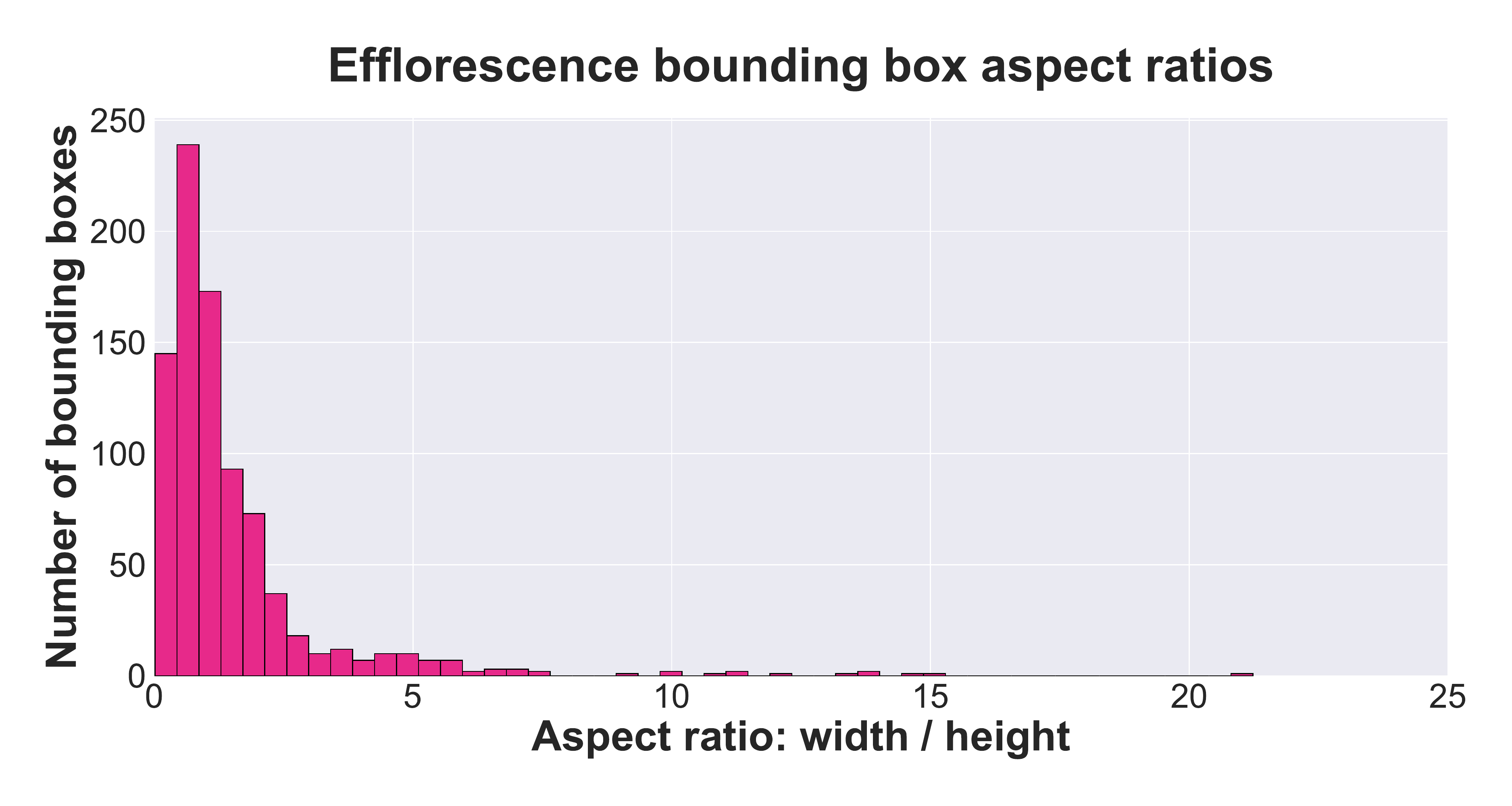} \\
	\includegraphics[width=0.33 \textwidth]{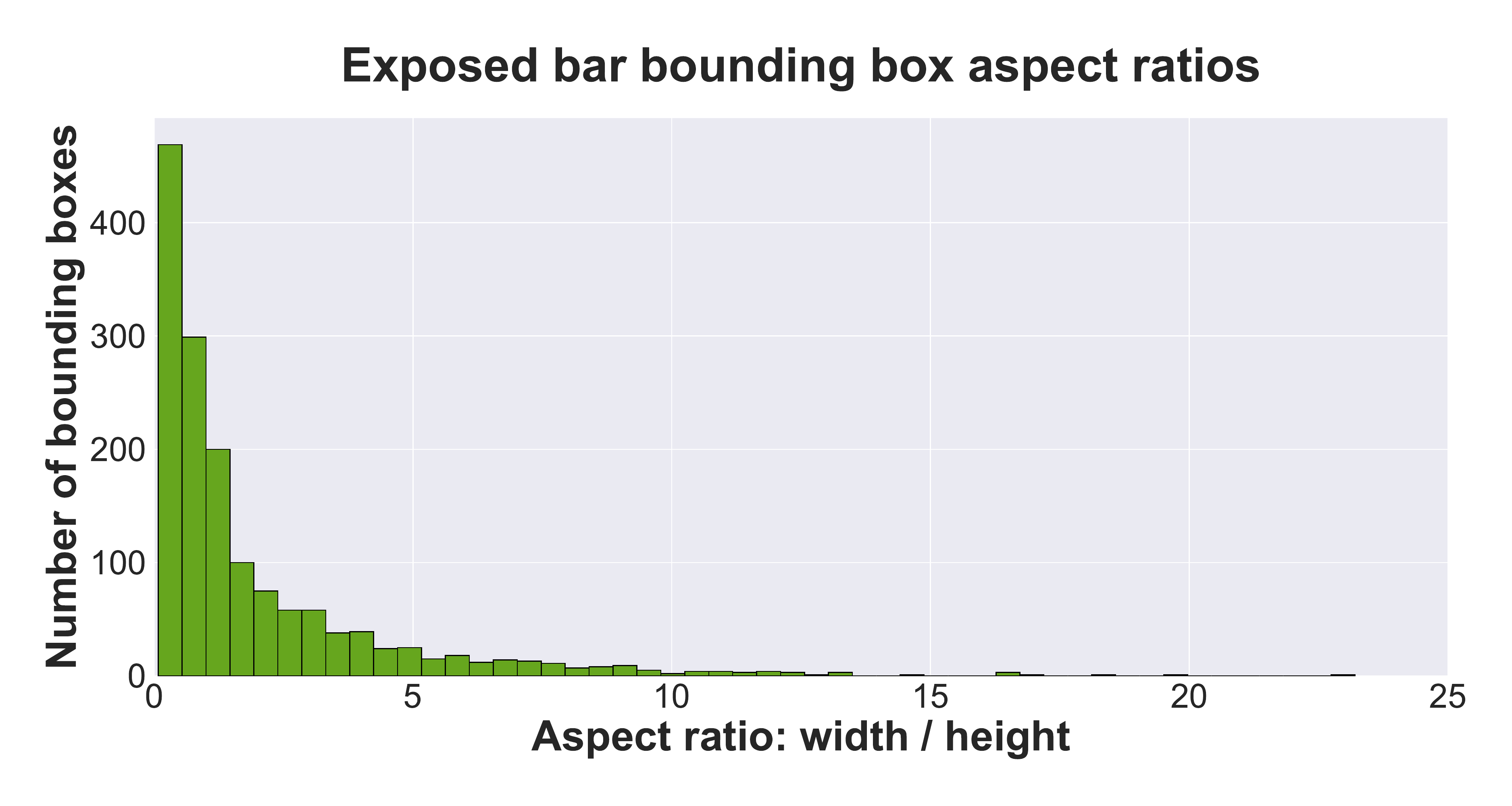}
	\includegraphics[width=0.33 \textwidth]{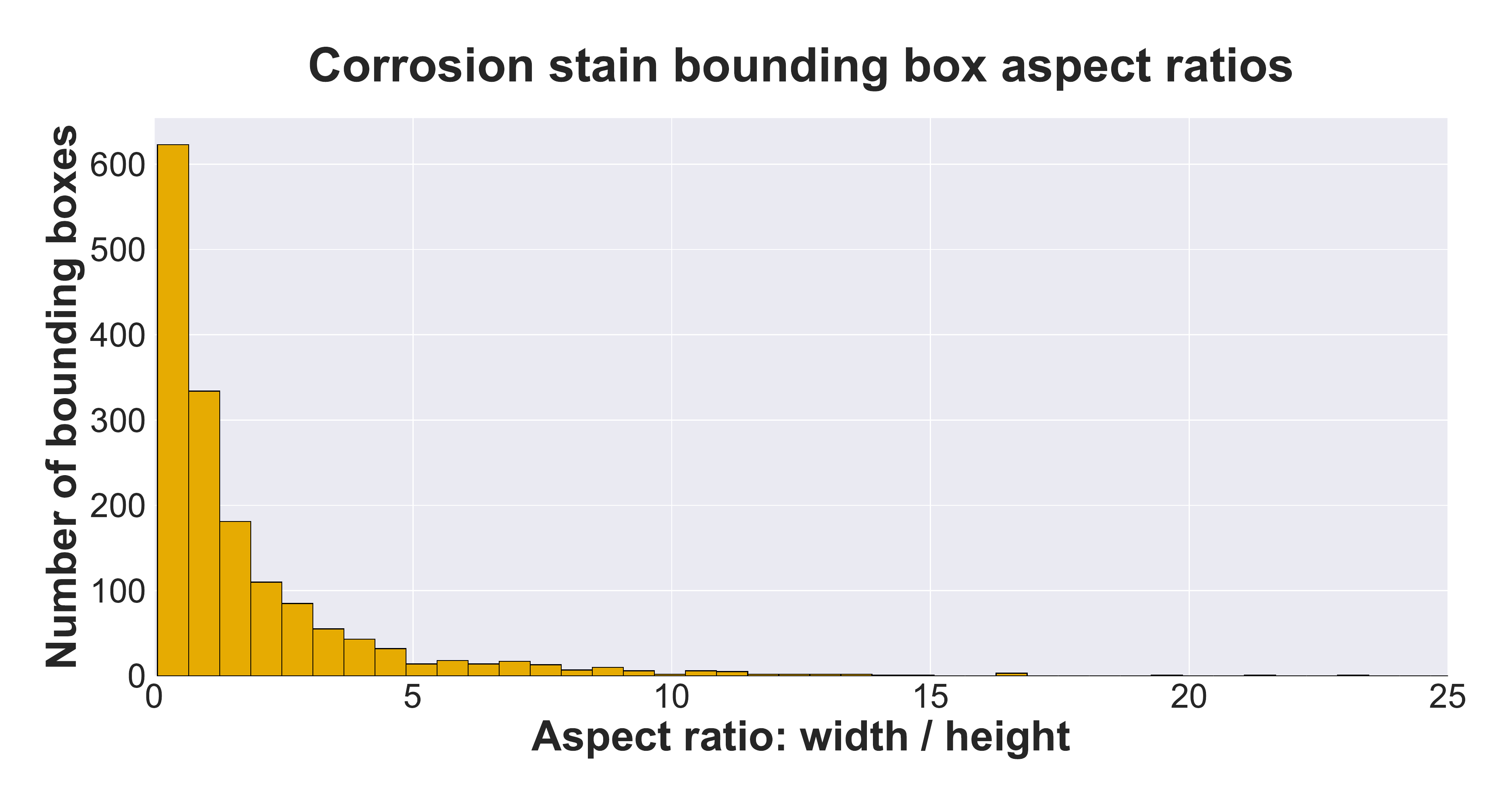}
	
	\caption{Individual distributions of number of bounding box annotations for different aspect ratios for each of the 5 defect classes.}
	\label{fig:aspect_ratios_per_defect}
\end{figure*}
We show additional information and statistics of the dataset. In figure \ref{fig:annotations} we show a histogram that demonstrates how one bounding box annotation typically contains more than one defect class at a time. In figure \ref{fig:annotations_per_image} the complementary histogram of the number of annotated bounding boxes per image can be found. Here, we can further observe that our choice of bridges led to image acquisition scenarios where one acquired image generally contains multiple different defect locations. While this is not impacting our classification task, we believe it is a crucial precursor for future extensions to a realistic semantic segmentation scenario.

In addition to the distribution of the annotated bounding box sizes for background and for all the defects combined as shown in the main body, the reader might be interested in the specific distribution per defect class. In figure \ref{fig:sizes_per_defect} the corresponding distribution of annotated bounding boxes per-class is shown. Similarly, figure \ref{fig:aspect_ratios_per_defect} contrasts the aspect ratio distributions for the individual defects. It is to be noted that these per-class distributions are not mutually exclusive because of multi-target overlap in the bounding box annotations. All individual classes have a similarly distributed bounding box size per defect including a long tail towards large resolutions. A major difference for individual classes can be found at high resolutions between the crack and efflorescence classes and the spallation, exposed bar and corrosion stain classes. The latter sometimes span an entire image. While this of course depends on the acquisition distance, we point out that in images acquired at a similar distance spalled and corroded areas including bar exposition are larger on average. 

\subsection{Random generation of background bounding boxes}
\begin{figure}[h!]
	\includegraphics[width=\columnwidth]{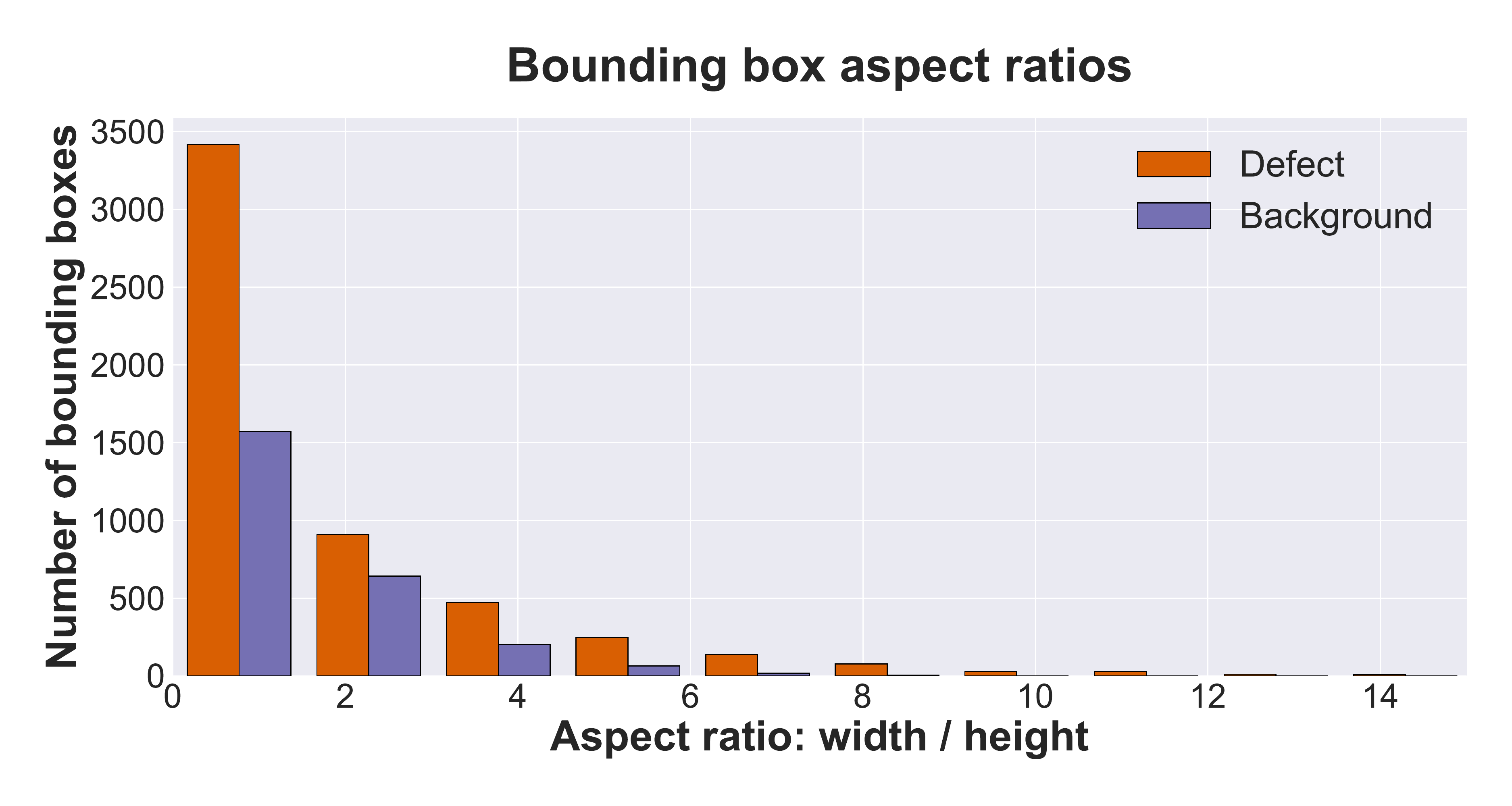}
	\caption{Distribution of number of defect and background bounding box annotations for different aspect ratios.}
	\label{fig:aspect_ratios}
\end{figure}
We emphasize that the CODEBRIM dataset has many factors that add to the complexity. Acquired images have large variations depending on the target geometry, types of defects and their overlapping behavior, the camera pose relative to the photographed surface (particularly if captured by UAV), as well as global scene properties such as illumination. From a machine learning classification dataset point of view it is thus interesting to capture this complexity in the generation of image patches for the background class. 

We therefore devote this supplementary section to provide further details to the reader on generation of background bounding boxes. Before administering the final dataset, the last dataset creation step of sampling areas containing background has been validated. Specifically, we have checked whether the distribution of sizes (shown in the main body) as well as the distribution of sampled areas' aspect ratios approximately follow those of the human annotated defects. In figure \ref{fig:aspect_ratios} we show the aspect ratios for the annotated defects together with the sampled bounding box aspect ratios for background. Whereas the overall count for background is less than the integrated total amount of defects (number of background samples roughly corresponds to occurrence of each individual defect class), the distribution of aspect ratios is confirmed to have the same trend. We have further made sure that none of the background bounding boxes have any overlap with bounding boxes annotated for defects and that bounding boxes for background are evenly distributed among images. In summary, this methodology captures the complexity of surface variations, target geometry, global illumination and makes sure that image patch resolution and sizes reflect those of defect annotations. 

\section{Deep convolutional neural networks for multi-target defect classification}\label{supp:dl}

\subsection{Per-class and multi-target CNN accuracies}
As mentioned in the main body of this work, most image classification tasks focus on the single target scenario and an easy pitfall would be to treat our task in a similar fashion. This would imply reporting classification accuracies independently per class and not treating the task in the multi-target fashion. We remind the reader that this would not represent the real-world scenario appropriately, where one is interested in the severity of the degradation of the inspected concrete structure. This severity is magnified if two or more different defect classes are mutually occurring and overlapping. Nevertheless, one idea could be to design the reward for the meta-learning algorithms based on such individual class accuracies or the corresponding average. We report the validation accuracy per class (background and five defects) and their respective average for the CNN literature baselines, together with the multi-target best validation accuracy and the corresponding test accuracy in table \ref{tab:baselines_per_class}. Note that we do this only for the sake of completeness as this thought could occur to other researchers and to show researchers the relationship between accuracy values. Initially, a multi-target accuracy of $65 \%$ might not look like a large value, but it practically translates to around $90 \%$ classification accuracy had each class been treated independently in our task. Apart from the above stated obvious argument of resemblance to real-world application, the table further indicates why the multi-target accuracy is a better metric to employ in meta-learning reward design. Although each of the baseline models learns to recognize individual defects in the image with high precision, they are not equally competent at recognizing all the defects together in the multi-target scenario. The individual class accuracies do not have clear trends as they fluctuate individually, are difficult to interpret from one model to the next and do not intuitively correlate with multi-target values. It is thus a bad idea to base evaluation and model comparison on single-target values and then later report multi-target accuracies as the former does not linearly correlate with the latter. We have noticed that rewards designed on the average per-class instead of multi-target accuracies lead to models that learn to predict only a subset of classes correctly and neglect overlaps as there is no reward for higher recognition rate of these overlaps. 

\begin{table*}[t]
\resizebox{\textwidth}{!}{\begin{tabular}{lrrrrrrrrr}
	Architecture  & \multicolumn{8}{c}{Accuracy [\%]}\\
	\cmidrule{2-10}
	 & mt best val & mt bv-test & bv-bg & bv-cr & bv-sp & bv-ef & bv-eb & bv-cs & bv-avg \\ 
	\toprule
	Alexnet & 63.05 & 66.98 & 89.30 & 89.30 & 89.93 & 90.72 & 93.71 & 88.05 & 90.16\\
	T-CNN & 64.30 & 67.93 & 90.09 & 87.89 & 89.62 & 88.99 & 94.49 & 87.57 & 89.77 \\
	VGG-A & 64.93 & 70.45 & 91.35 & 90.25 & 89.93 & 90.56 & 93.55 & 86.47 & 90.35 \\
	VGG-D & 64.00 & 70.61 & 90.72 & 91.82 & 89.93 & 89.30 & 93.71 & 87.42 & 90.48\\
	WRN-28-4 & 52.51 & 57.19 & 87.89 & 84.11 & 85.53 & 84.43 & 89.15 & 80.34 & 85.24\\
	DenseNet-121 & 65.56 & 70.77 & 91.51 & 89.62 & 87.75 & 89.10 & 94.49 & 87.73 & 90.03\\
	\bottomrule

\end{tabular}}
\caption{Best validation model's accuracies for each individual class (bg - background, cr - crack, sp - spallation, ef - efflorescence, eb - exposed bars, cs - corrosion stain) and their average (avg) shown together with the multi-target accuracy (mt best val) and the corresponding multi-target test accuracy (mt bv-test).}
\label{tab:baselines_per_class}
\end{table*}

\begin{table*}[t]
\resizebox{\textwidth}{!}{\begin{tabular}{l|l|l|l}
\toprule
 Layer type & MetaQNN-1 & MetaQNN-2 & MetaQNN-3\\ 
  \bottomrule
  conv 1 & $9\times9$ - 256, s = 2 & $5\times5$ - 128 & $3\times3$ -  128, p = 1;\quad $1\times1$ - 128 (skip to 3)\\
  conv 2 & $3\times3$ - 32, p = 1 & $7\times7$ - 32, s = 2 &  $3\times3$ - 128, p = 1\\
  conv 3 & $5\times5$ - 256 & $3\times3$ -  256, p = 1;\quad $1\times1$ - 256 (skip to 5)& $9\times9$ -  128, s = 2\\
  conv 4 & $7\times7$ - 256, s = 2 & $3\times3$ - 256, p = 1 & $3\times3$ -  256, p = 1;\quad $1\times1$ - 256 (skip to SPP)\\
  conv 5 & & $3\times3$ - 32& $3\times3$ - 256, p = 1\\
  conv 6 & & $9\times9$ - 128, s = 2& \\
  \midrule
  SPP & scales = 4 & scales = 3 & scales = 4\\
  FC 1 & 128 & 128 & 64\\
  classifier & linear - 6, sigmoid & linear - 6, sigmoid & linear - 6, sigmoid\\
  \bottomrule
\end{tabular}}
\caption{Top three neural architectures of MetaQNN for our task. Convolutional layers (conv) are parametrized by a quadratic filter size followed by the amount of filters. p and s represent padding and stride respectively. If no padding or stride is specified then $p = 0$ and $s = 1$. Skip connections are an additional operation at a layer, with the layer where the connection is attached to specified in brackets. A spatial pyramidal pooling (SPP) layer connects the convolutional feature extractor part to the classifier. Every convolutional and FC layer are followed by a batch-normalization and a ReLU and each model ends with a linear transformation with a Sigmoid function for multi-target classification.}
\label{tab:MetaQNN_ArcDefinitions}
\end{table*}

\subsection{Meta learned architecture definitions}
We show the detailed configurations of the top three MetaQNN and ENAS neural architectures for which accuracies are shown in the main body. 

Table \ref{tab:MetaQNN_ArcDefinitions} shows the definitions for the top three meta-learned models from MetaQNN on our task. Each convolutional layer is expressed through quadratic filter size and number of filters, followed by an optional specification of padding or stride. If a skip connection/convolution has been added it is added as an additional operation on the same level and we specify the layer to which it skips to. The SPP layer is characterized by the number of scales at which it pools its feature input. As an example, scales = $4$ indicates four adaptive pooling operations such that the output width times height corresponds to $1 \times 1, 2 \times 2, \ldots 4 \times 4$. The fully-connected (FC) layer is defined by the number of feature outputs it produces. All convolutional and FC layers are followed by a batch-normalization and a ReLU layer. 

\begin{figure*}
	\centering
	\begin{subfigure}{0.475\textwidth}
		\includegraphics[width = \textwidth]{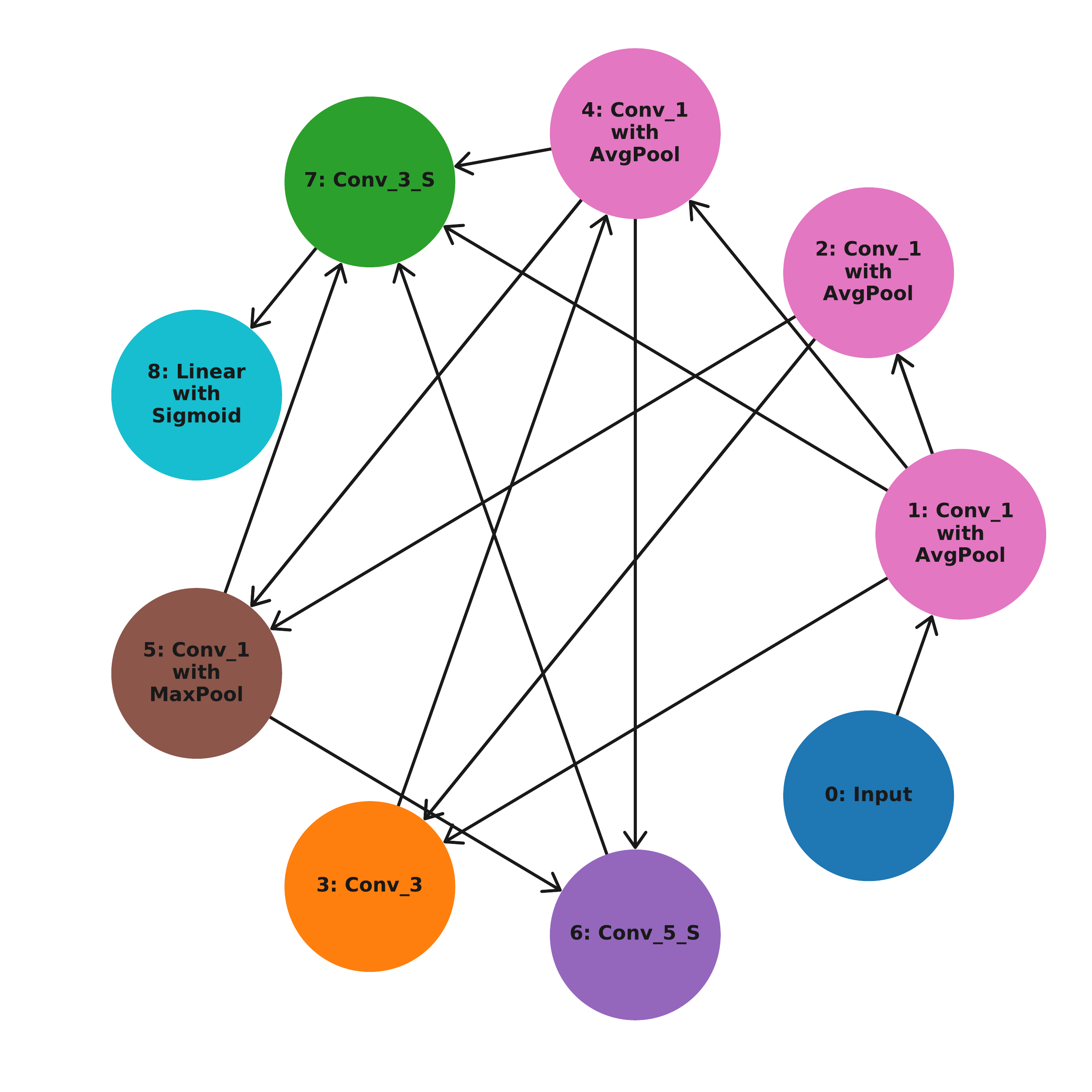}
		\caption{ENAS-1}
	\end{subfigure}
	\begin{subfigure}{0.475\textwidth}
		\includegraphics[width = \textwidth]{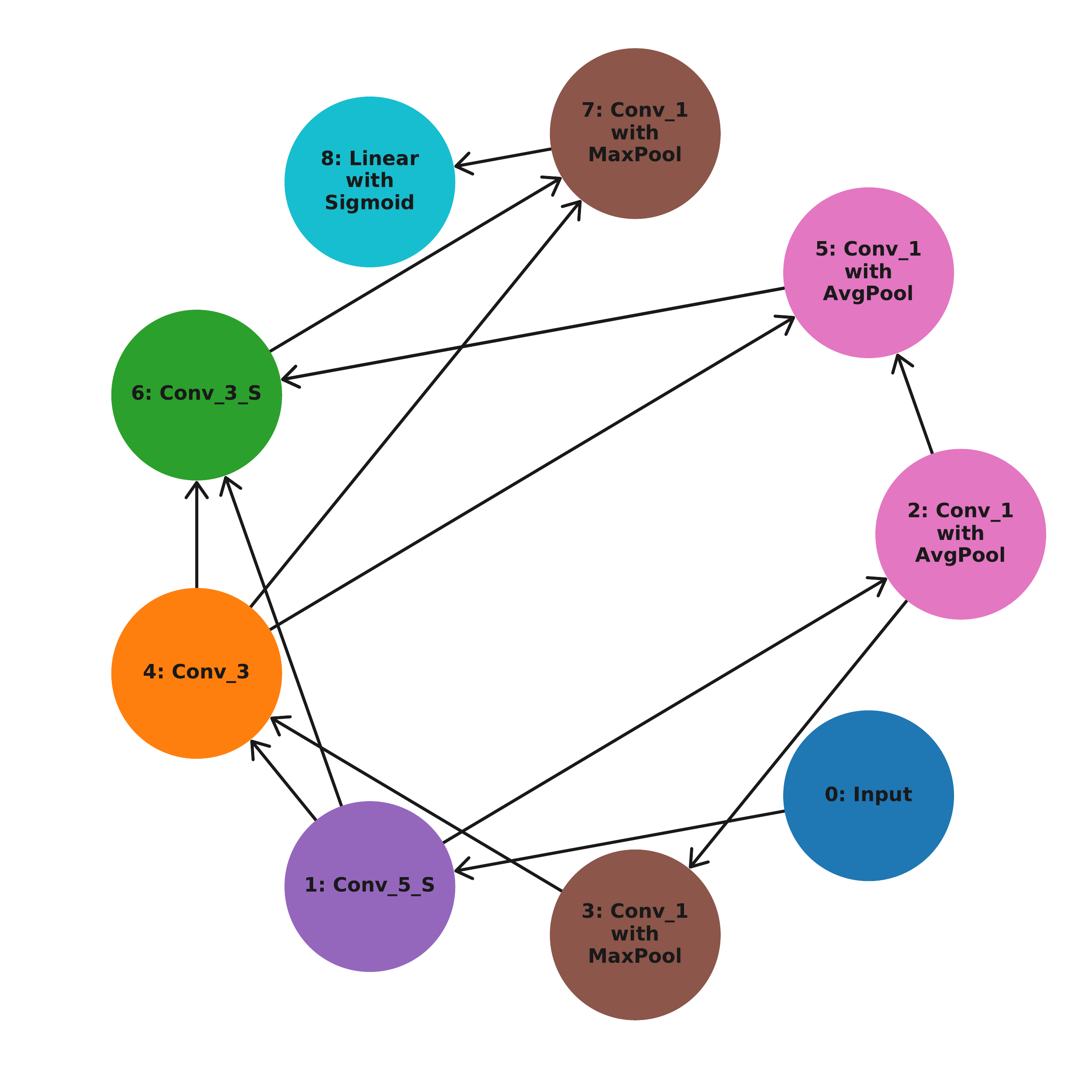}
		\caption{ENAS-2}
	\end{subfigure} 
	\\
	\begin{subfigure}{0.475\textwidth}
		\includegraphics[width = \textwidth]{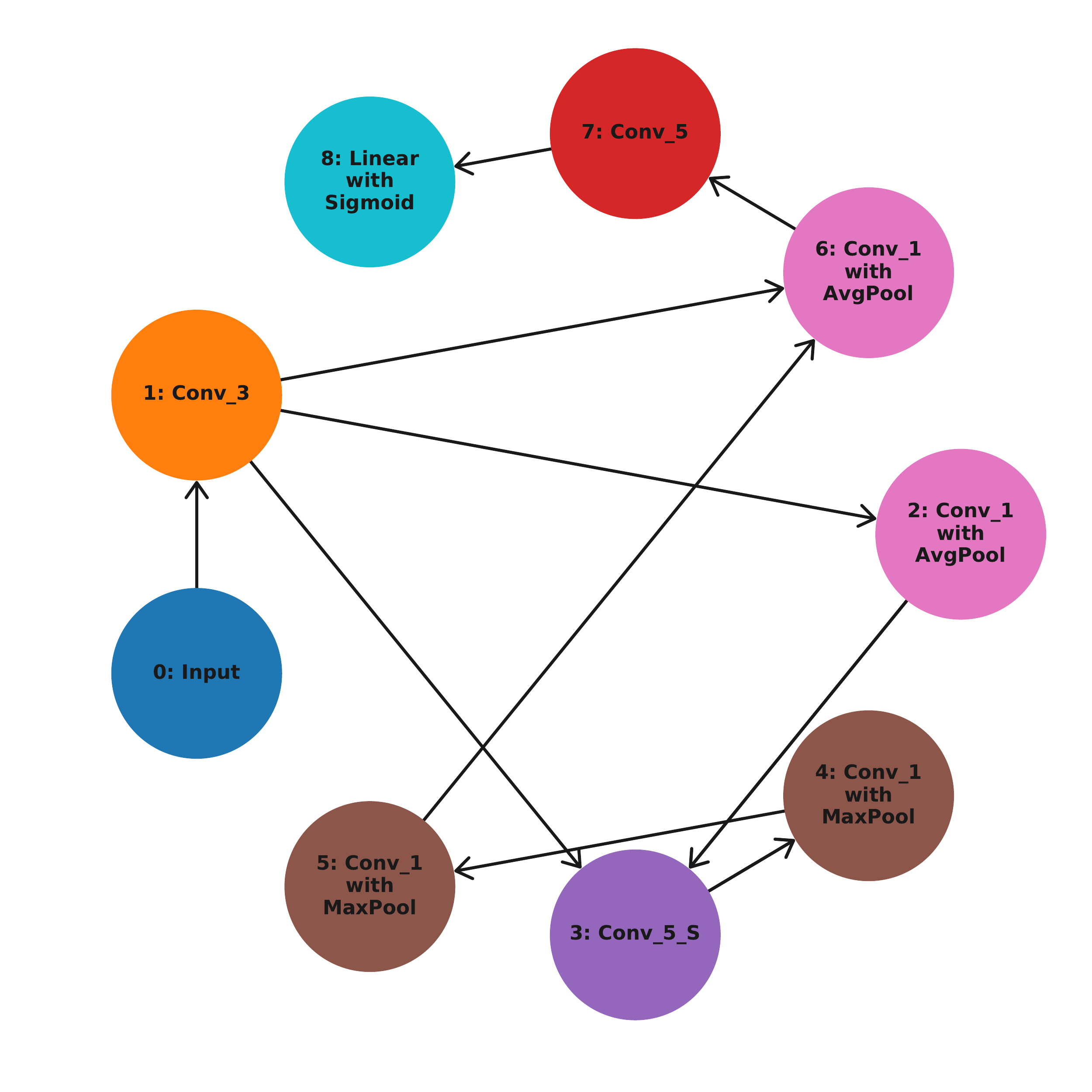}
		\caption{ENAS-3}
	\end{subfigure}
	\caption{Top three neural architectures of ENAS for our task. Convolutions (conv) are denoted with quadratic filter size and a post-fix "S" for depth-wise separability. MaxPool and AvgPool are max and average pooling stages with $3 \times 3$ windows. ENAS uses a pre-determined amount of features per convolutional layer during the search and during final training uses a growth strategy of $k=2$ similar to DenseNets. The amount of features per convolution is $64$, doubled by the growth strategy after layers $2$ and $4$. The graph is acyclic and all connections between layers are indicated by directed arrows. }
	\label{fig:ENAS_ArcDefinitions}
\end{figure*}

Figure \ref{fig:ENAS_ArcDefinitions} shows graphical representations of the top three neural models of ENAS for our task. All of the ENAS architectures have seven convolutional layers followed by a linear transformation as defined prior to the search. We have chosen a visual representation instead of a table because the neural architectures (acyclic graphs) contain many skip connections that are easier to perceive this way. All convolutions have quadratic filter size and a base amount of $64$ features that is doubled after the second and forth layer as defined by a DenseNet growth strategy with $k = 2$. 

\subsection{Transferring ImageNet and MINC features}
\begin{table}[h!]
\resizebox{\columnwidth}{!}{
\begin{tabular}{lrrr}
\multicolumn{4}{c}{\textbf{Transfer learning}} \\
Architecture &Source &  \multicolumn{2}{c}{Accuracy [\%]} \\
\cmidrule{3-4}
 &  & best val& bv-test\\ 
\toprule
Alexnet & ImageNet & 60.53 & 62.87  \\ 
VGG-A & ImageNet & 60.22 & 66.35  \\ 
VGG-D & ImageNet & 56.13 & 65.56  \\ 
Densenet-121 & ImageNet & 54.71 & 57.66 \\
Alexnet & MINC & 60.06 & 66.50  \\ 
VGG-D & MINC & 61.47 & 67.14 \\ 
\bottomrule 
\end{tabular}}
\caption{Multi-target best validation and best validation model's test accuracy for fine-tuned CNNs with convolutional feature transfer from models pre-trained on the MINC and ImageNet datasets.}
\label{tab:transferlearning}
\end{table}
We investigate transfer learning with features pre-trained on the ImageNet and MINC datasets for a variety of neural architectures by using pre-trained weights provided by corresponding original authors. We fine-tune these models by keeping the convolutional features constant and only training the classification stage for 70 epochs with a cycled learning rate and other hyper-parameters as specified in the main body. Best multi-target validation and associated test values are reported in table \ref{tab:transferlearning}. Although the pre-trained networks initially train much faster, we observe that transferring features from the unrelated ImageNet and MINC tasks does not help, it in fact hinders the multi-target defect classification task. We postulate that this could be due to a variety of factors like the task being too unrelated with respect to the combination of object and texture recognition demanded by our task. This observation matches previous work investigating transfer learning of object related features to texture recognition problems. In such a scenario, the authors of \cite{Zhang2017} find the need to evaluate feature importance and selectively integrate only a subset of relevant ImageNet object features to yield performance benefits for texture recognition and prevent performance degradation. We further hypothesize another possibility that the multi-target property of the task could require a different abstraction of features from those already present in the convolutional feature encoder of the pre-trained models. Further investigation of transfer learning should thus consider an approach that doesn't include all pre-trained features, selects a subset of pre-trained weights or employs different fine-tuning strategies.

\subsection{Classification examples}
In addition to the accuracy values reported in the main body, we show qualitative example multi-target classifications as predicted by our trained MetaQNN-1 model. We do this to give the reader a more comprehensive qualitative understanding of the complexity and challenges of our multi-target dataset. In order to better outline these challenges, we separate these examples into the following three categories:

\begin{enumerate}
	\item Correct multi-target classification examples where all labels are predicted correctly.
	\item False multi-target classification examples where at least one present defect class is recognized correctly, but one or more defect classes is missed or falsely predicted in addition. 
	\item False multi-target classification examples where none of the present defect classes is recognized correctly. 
\end{enumerate}

\begin{figure*}[t]
	\begin{subfigure}{\textwidth}
		\includegraphics[width = 0.16\textwidth]{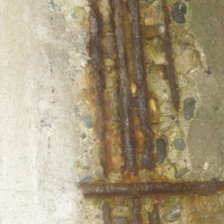}
		\includegraphics[width = 0.16\textwidth]{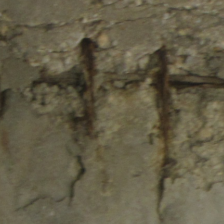}
		\includegraphics[width = 0.16\textwidth]{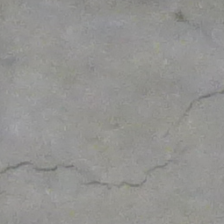}
		\includegraphics[width = 0.16\textwidth]{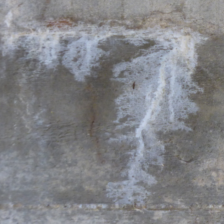}
		\includegraphics[width = 0.16\textwidth]{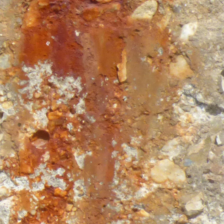}
		\includegraphics[width = 0.16\textwidth]{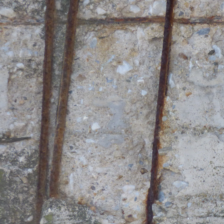}
		\caption{Correct multi-target classification examples from the validation set. From left to right: 1.) exposed bar, corrosion, spalling. 2.) spallation, exposed bars, corrosion and cracks. 3.)  crack. 4.) efflorescence 5.) spallation and corrosion. 6.) spallation with exposed bars. \\}
	\end{subfigure}
	
	\begin{subfigure}{\textwidth}
		\includegraphics[width = 0.16\textwidth]{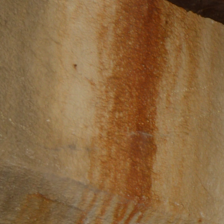}
		\includegraphics[width = 0.16\textwidth]{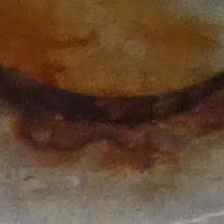}
		\includegraphics[width = 0.16\textwidth]{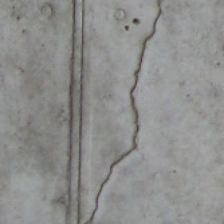}
		\includegraphics[width = 0.16\textwidth]{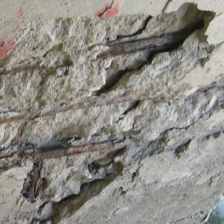}
		\includegraphics[width = 0.16\textwidth]{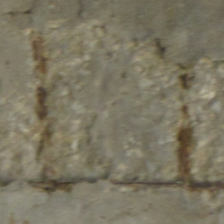}
		\includegraphics[width = 0.16\textwidth]{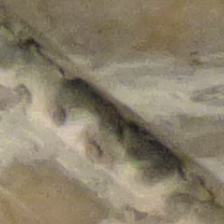}
		\caption{False multi-target classification examples from the validation set where at least one present defect class is recognized correctly. From left to right: 1.) corrosion (predicted corrosion and efflorescence). 2.) corrosion (predicted corrosion, spallation and exposed bar). 3.) crack (predicted crack and efflorescence). 4.) spallation, exposed bar, corrosion (predicted spallation and corrosion). 5.) spallation, exposed bar, corrosion (predicted crack and corrosion). 6.) efflorescence (predicted efflorescence and crack). \\}
	\end{subfigure}
	
	\begin{subfigure}{\textwidth}
		\includegraphics[width = 0.16\textwidth]{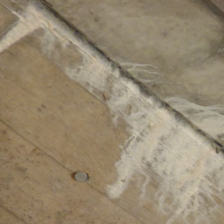}
		\includegraphics[width = 0.16\textwidth]{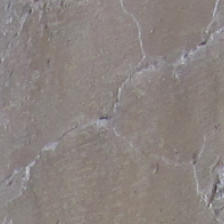}
		\includegraphics[width = 0.16\textwidth]{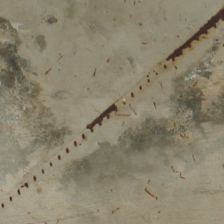}
		\includegraphics[width = 0.16\textwidth]{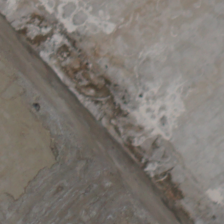}
		\includegraphics[width = 0.16\textwidth]{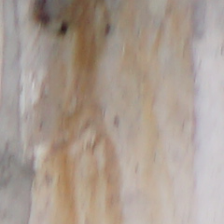}
		\includegraphics[width = 0.16\textwidth]{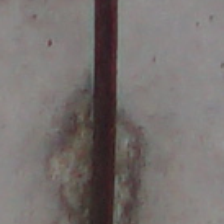}
		\caption{False multi-target classification examples from the validation set where none of the present defect categories is recognized correctly. From left to right: 1.) efflorescence (predicted background). 2.) crack (predicted background). 3.) exposed bar with corrosion (predicted background). 4.) efflorescence (predicted background). 5.) corrosion (predicted spallation). 6.) exposed bar (predicted crack).}
	\end{subfigure}
	\caption{Multi-target classification examples from the validation set using the trained MetaQNN-1 model.}
	\label{fig:classifications}
\end{figure*}

Corresponding images, together with ground-truth labels and the model's predictions are illustrated in the respective parts of figure \ref{fig:classifications}. The few shown examples were picked to show the variety of different defect types and their combinations. Overall, the images show the challenging nature of the multi-target task. Whereas the majority of multi-target predictions are correct, the trained models face a number of different factors that make classification difficult. Particularly, partially visible defect classes, amount of overlap, variations in the surface, different exposure and illumination can lead to the model making false multi-target predictions, where only a subset of targets is predicted correctly.


\subsection{Alternative dataset splits}
\begin{table}[h!]
\resizebox{\columnwidth}{!}{
\begin{tabular}{lrrrr}
Architecture & \multicolumn{2}{c}{Multi-target accuracy [\%]} & Params [M] & Layers\\
\cmidrule{2-3}
 & val & test & & \\ 
\toprule
Alexnet & 63.50 & 62.94 & 57.02 & 8\\ 
T-CNN & 63.87 & 63.00 & 58.60 & 8\\ 
VGG-A & 65.33 & 61.93 & 128.79 & 11\\ 
VGG-D & 63.76 & 62.50 & 134.28 & 16\\ 
WRN-28-4  & 59.75 & 55.56 & 5.84 & 28\\ 
Densenet-121 & 66.54 & 65.93 & 11.50 & 121\\
\midrule
ENAS-1 & 67.71 & 66.31 &  3.41 & 8\\ 
ENAS-2 & 66.50 & 64.37  & 2.71 & 8\\ 
ENAS-3 & 65.66 & 65.81 & 1.70 & 8\\ 
MetaQNN-1 & 66.70 & 65.91  & 4.53 & 6 \\ 
MetaQNN-2 & 65.25 & 64.82  & 1.22 & 8\\ 
MetaQNN-3 & 70.95 & 67.56 & 2.88 & 7\\
\bottomrule
\end{tabular} }$\smallskip$
\caption{Evaluation in analogy to table 2 of the main body, but on alternative dataset splits based on a per-bridge separation.}
\label{tab:performances}
\end{table}
The final dataset presented in the main portion of the paper has been chosen to contain a random set of 150 unique defect examples per class for validation and test sets respectively. To avoid over-fitting we have further added the constraint that all crops stemming from bounding boxes from one image must be contained in only one of the dataset splits. The rationale behind this choice is to ensure a non-overlapping balanced test and validation set in order to avoid biased training that favors certain classes and report skewed loss and accuracy metrics. 

A different alternative way of conducting such a validation and test split is to split the data based on unique bridges. Such an approach however features multiple challenges that make it infeasible to apply in practice. In particular, not every bridge has the same amount of defects and not every bridge has the same amount of defects per class. Typically also defects of varying severity and overlap are featured (e.g. some have more early-stage cracks than exposed bars). The main challenges thus are:
\begin{enumerate}
\item Only a certain combination of unique bridges can yield an even approximately balanced dataset split in terms of class presence.
\item Creation of class-balanced splits relies on either excluding some of the highly occurring defects or leaving the dataset split only approximately balanced. The latter could result in training a model that favors a particular class and skewed average metrics being reported. The former can result in omitting particularly challenging or easy instances from the validation or test set and accordingly distorting the interpretation of the model's accuracy.  
\item Even when balancing the classes approximately by choosing complementary bridges, the severity of defects is not necessarily well sampled or balanced. 
\end{enumerate}
On the other hand, a bridge-based dataset split provides more insights with respect to over-fitting concrete properties such as surface roughness, color, context or, given that images at different bridges were acquired at different points in time with variations in global scene conditions. We therefore nevertheless investigate an alternative bridge-based dataset split that is based on three bridges for validation and test set respectively. The bridges have been chosen such that the resulting splits are approximately balanced in terms of class occurrence, albeit with the crack category being more present and the efflorescence class being under-sampled. The resulting accuracies should thus be considered with caution in direct comparison to the main paper. 

Using this alternate dataset split we retrain all neural architectures presented in the main paper. We note that we have not repeated the previous hyper-parameter grid-search and simply use the previously obtained best set of hyper-parameters. In analogy, the meta-learning architecture search algorithms have not been used to sample new architectures specific to this dataset variant. The obtained final validation and test accuracies are reported in table \ref{tab:performances}. We re-iterate, that although we have coined the splits validation and test set, the sets can be used interchangeably here as no hyper-parameter tuning has been conducted on the validation set. 

\noindent Obtained accuracies are similar to the experimental results presented in the paper's main body. We can observe that meta-learned architectures are not in exact previous order, e.g. MetaQNN3 outperforms MetaQNN1. However, meta-learned architectures still outperform the baselines and previous conclusions therefore hold. Due to the previously presented challenges in creation of an unbiased bridge-based dataset we therefore believe our dataset splits presented in the main body to be more meaningful to assess the models' generalization capabilities. 

%

\end{document}